\documentclass[10pt,twocolumn,letterpaper]{article}

 \usepackage[pagenumbers]{cvpr} 

\newlength\savewidth\newcommand\shline{\noalign{\global\savewidth\arrayrulewidth
  \global\arrayrulewidth 1pt}\hline\noalign{\global\arrayrulewidth\savewidth}}
\newcommand{\tablestyle}[2]{\setlength{\tabcolsep}{#1}\renewcommand{\arraystretch}{#2}\centering\footnotesize}

\usepackage{array}
\newcolumntype{x}[1]{>{\centering\arraybackslash}p{#1pt}}
\usepackage{booktabs}
\usepackage{arydshln}

\definecolor{PastelGreen}{rgb}{0.13, 0.55, 0.13}%
\newcommand{\emphhigh}[1]{\textcolor{ForestGreen}{#1}}

\definecolor{PastelRed}{rgb}{0.8, 0.13, 0.13}    %

\definecolor{cvprblue}{rgb}{0.21,0.49,0.74}
\usepackage[pagebackref,breaklinks,colorlinks,allcolors=cvprblue]{hyperref}
\usepackage{url}
\usepackage{amsfonts, amsmath, amsthm}       
\usepackage{nicefrac}       
\usepackage{microtype}     
\usepackage[symbol]{footmisc}        
\usepackage{paralist}
\usepackage{arydshln}
\usepackage{comment}
\usepackage[linesnumbered,ruled,vlined]{algorithm2e}
\usepackage{array, booktabs, multirow, makecell, float} 
\usepackage{enumitem}
\usepackage[table]{xcolor}  
\usepackage{array} 

\definecolor{KleinBlue}{rgb}{0.0, 0.129, 0.6}

\newif\ifprintcomments
\printcommentsfalse  

\usepackage{lipsum}

\usepackage{listings}
\usepackage{caption}
\usepackage[utf8]{inputenc}

\lstdefinestyle{pytorch}{
  language=Python,
  basicstyle=\ttfamily\scriptsize,
  keywordstyle=\bfseries,
  commentstyle=\color{teal},
  numberstyle=\tiny\color{gray},
  numbers=none,
  frame=tb,                
  rulecolor=\color{black}, 
  framerule=1pt,           
  showstringspaces=false,
  tabsize=2,
  breaklines=true,
  breakatwhitespace=true,
  columns=fullflexible,
  keepspaces=true,
  aboveskip=1pt,
  belowskip=5pt,
}

\newcommand{\TODO}[1]{}

\newtheorem{statement}{Statement}

\definecolor{Gray}{gray}{0.85}
\definecolor{ForestGreen}{rgb}{0.13, 0.55, 0.13}
\newcolumntype{a}{>{\columncolor{Gray}}c}

\def\confName{CVPR}
\def\confYear{2026}

\title{SpaceVLM: Sub-Space Modeling of Negation in Vision-Language Models}

\author{
Sepehr Kazemi Ranjbar\thanks{Equal contribution.} \\
Independent Researcher \\
{\tt\small sepehrkazemi9@gmail.com}
\and
Kumail Alhamoud\footnotemark[1] \\
MIT \\
{\tt\small kumail@mit.edu}
\and
Marzyeh Ghassemi \\
MIT \\
{\tt\small mghassem@mit.edu}
}

\begin{document}
\maketitle
\begin{abstract}
Vision-Language Models (VLMs) struggle with negation. Given a prompt like “retrieve (or generate) a street scene without pedestrians,” they often fail to respect the “not.” Existing methods address this limitation by fine-tuning on large negation datasets, but such retraining often compromises the model’s zero-shot performance on affirmative prompts. We show that the embedding space of VLMs, such as CLIP, can be divided into semantically consistent subspaces. Based on this property, we propose a training-free framework that models negation as a subspace in the joint embedding space rather than a single point (\Cref{fig:intro}). To find the matching image for a caption such as “A but not N,” we construct two spherical caps around the embeddings of A and N, and we score images by the central direction of the region that is close to A and far from N. Across retrieval, MCQ, and text-to-image tasks, our method improves negation understanding by about $30\%$ on average over prior methods. It closes the gap between affirmative and negated prompts while preserving the zero-shot performance that fine-tuned models fail to maintain. Code will be released upon publication.
\end{abstract}    
\vspace{-5pt}
\section{Introduction}
\label{sec:intro}
Joint embedding-based Vision-Language Models (VLMs) \cite{radford2021learning, li2022blip, zhai2023sigmoid}, such as CLIP \cite{radford2021learning}, have become strong foundations for visual understanding. These models consist of an image encoder and a text encoder that map visual and textual inputs into a shared embedding space, where similarity is measured by dot product. When pretrained on massive image–text datasets, they exhibit strong generalization and are widely used for classification, retrieval, and text-to-image generation \cite{zhang2024vision, zhao2023clip, klemmer2025satclip}, with successful applications in specialized domains such as healthcare \cite{lu2024visual}. However, they struggle with inputs that require logical reasoning \cite{kamath-etal-2023-text, lewis2022does, rasekh2025multi}, particularly those involving negation \cite{alhamoud2025vision, singh2025learning}.

Consider the query ``retrieve an image with a dog but not a cat.'' A model processing this input must correctly exclude images containing cats, while retaining valid alternatives that include dogs. Yet, as shown by prior work \cite{alhamoud2025vision}, CLIP-like models \cite{radford2021learning, zhai2023sigmoid, yuksekgonul2022and, singh2025learning} fail to interpret negation in their standard inference setup. Previous studies attributed this weakness to the lack of negation-rich captions in the training data; to address this, they generated large synthetic datasets and fine-tuned VLMs on negation-enriched image–text pairs \cite{yuksekgonul2022and, singh2025learning, park2025know}. Yet these fine-tuning methods face two limitations: (i) they fail to fully close the performance gap between affirmative and negated queries, and (ii) they often reduce the model’s zero-shot generalization on tasks unrelated to negation. This raises a central question: \emph{can negation be modeled effectively without any fine-tuning?}

\begin{figure}
    \centering
    \includegraphics[width=\linewidth, height=3.5cm]{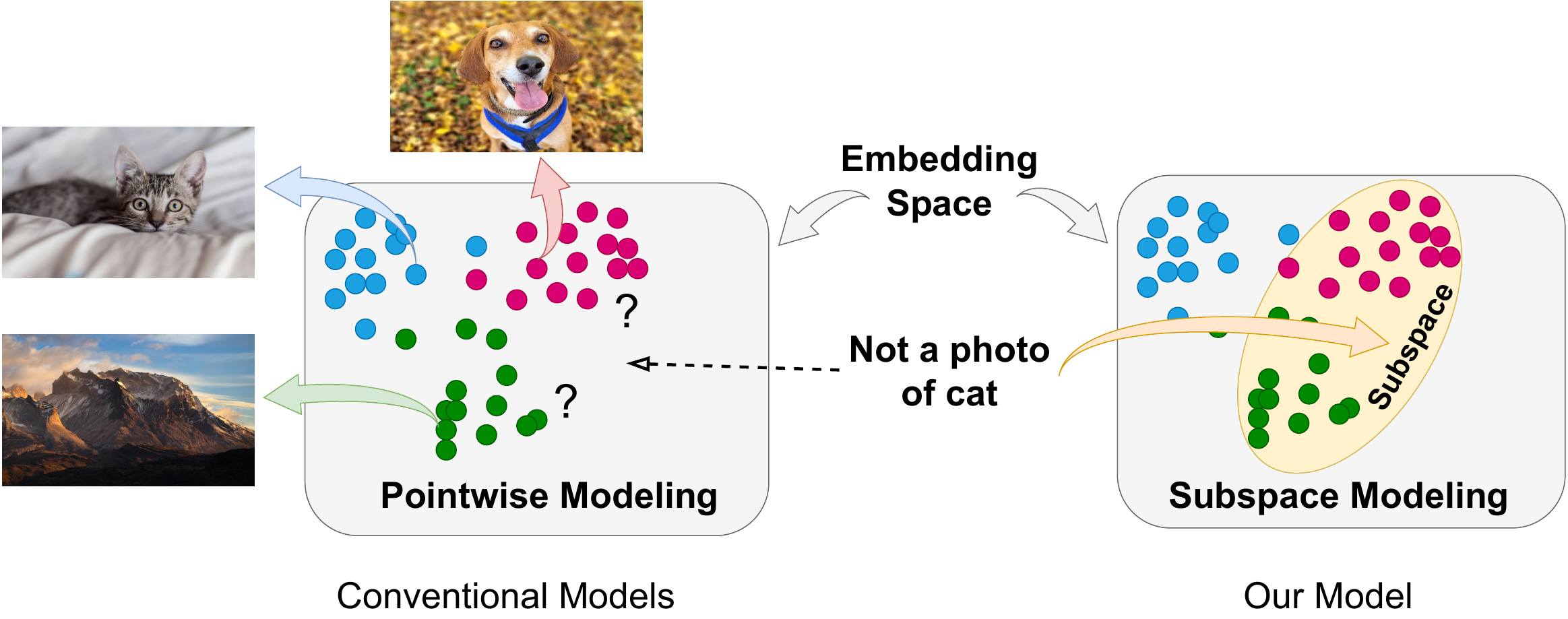}
    \caption{Given a caption such as \texttt{``Not a photo of a cat''}, standard VLM approaches attempt to map this negative caption to a single point in the embedding space, which makes it ambiguous where the correct destination should be. In contrast, our approach maps the negative caption to a subspace rather than a point, enhancing the model’s ability to handle negation effectively.}
    \label{fig:intro}
\end{figure}

\begin{figure*}[ht]
\centering
\includegraphics[width=\textwidth]{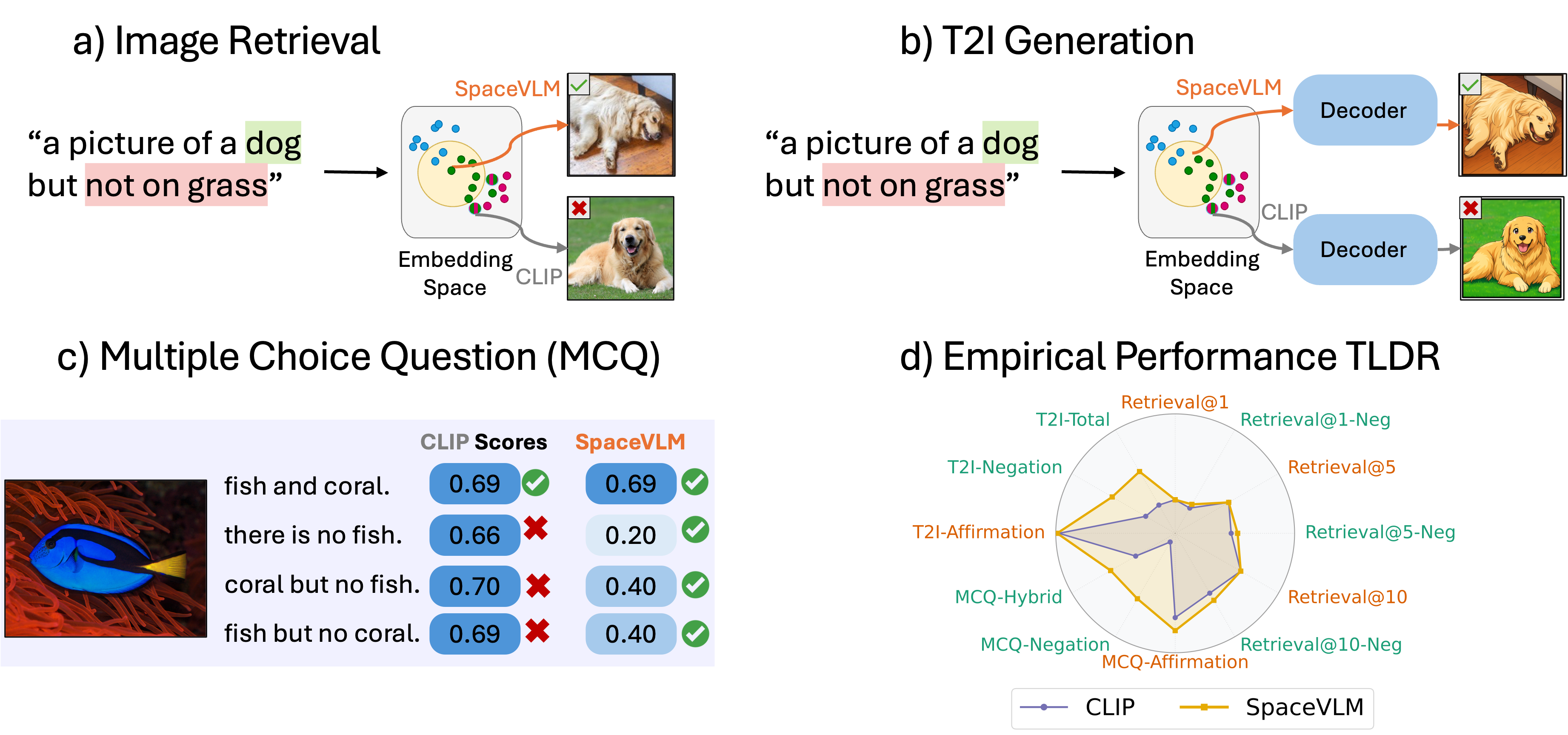}
\caption{
In both (a) Image Retrieval and (b) Text-to-Image (T2I) Generation, \textbf{\textcolor{gray}{CLIP}} embeds the input prompt “a picture of a \textcolor{ForestGreen}{dog} but \textcolor{BrickRed}{not on grass}” near images that include both \textcolor{ForestGreen}{dog} and \textcolor{BrickRed}{grass}, leading to incorrect retrievals or generations. In (c) MCQ, \textbf{\textcolor{gray}{CLIP}} assigns similar similarity scores to all captions mentioning “fish” and “coral,” regardless of whether they include or exclude a concept, leading to incorrect image-text matching. By modeling negation as a subspace, \textbf{\textcolor{Orange}{SpaceVLM}} fixes all these issues. As summarized in (d), this geometric modeling of \textbf{\textcolor{Orange}{SpaceVLM}} empirically improves negation understanding across these tasks, while preserving performance on affirmative prompts.
}
\vspace{-12pt}
   \label{fig:task_overview}
\end{figure*}

First, we motivate why fine-tuning alone cannot fully solve negation.
The key issue is that ``not a cat'' excludes cat, but leaves open many alternatives, such as dog or apple. Representing this with a single embedding vector, i.e., following the dot-product scoring used in joint embedding-based VLMs such as CLIP \cite{radford2021learning}, SigLIP \cite{zhai2023sigmoid}, or LiT-tuned AIMV2 \cite{zhai2022lit, fini2025multimodal}, is inherently insufficient. To account for infinitely many valid possibilities, negation cannot be modeled by a single point in the VLM embedding space (\Cref{fig:intro}). In contrast, we verify that CLIP’s embedding space can be divided into semantically consistent subspaces \cite{Bhalla2024InterpretingCW,zhao2025quantifying}. We then model negation as the intersection between an affirmative and a complementary \emph{subspace}, and derive a simple, training-free scoring rule. For a caption ``A but not N,'' we compute two spherical caps centered at the embeddings of A and N, and use the central direction of the region that is close to A and far from N to score images. Because this scoring operates purely at inference time, it leaves the model’s behavior unchanged on queries without negation, ensuring no degradation on unrelated tasks. Importantly, our SpaceVLM framework is model-agnostic and applicable to any joint embedding-based VLM.

We validate SpaceVLM across more than $40$ experimental settings spanning combinations of VLM backbones, image and video datasets, and diverse negation tasks including multimodal retrieval, Multiple Choice Question (MCQ), and text-to-image generation (\Cref{fig:task_overview}). Following NegBench~\cite{alhamoud2025vision}, we use the COCO~\cite{lin2014microsoft}, VOC-2007~\cite{pascal-voc-2007}, and MSR-VTT~\cite{xu2016msr} datasets for general-domain retrieval and MCQ, and CheXpert~\cite{irvin2019chexpert} for medical diagnostics with negation. Our training-free framework consistently improves negation understanding for every joint embedding–based model tested — CLIP~\cite{radford2021learning}, SigLIP~\cite{tschannen2025siglip}, NegCLIP~\cite{yuksekgonul2022and}, ConCLIP~\cite{singh2025learning}, AimV2~\cite{fini2025multimodal}, BiomedCLIP~\cite{zhang2023biomedclip}, and others — while preserving zero-shot performance on affirmative queries. Despite requiring no training or architectural modification, SpaceVLM outperforms fine-tuned baselines such as CLIP-NegFull~\cite{alhamoud2025vision}, ConCLIP~\cite{singh2025learning}, NegCLIP~\cite{yuksekgonul2022and}, and NegationCLIP~\cite{park2025know}, and it even surpasses the recent geometric approach DCSM~\cite{kang2025clip}. 

Ablation studies show that the cosine-similarity threshold, the main hyperparameter in SpaceVLM, is robust within a practical range, making it easy to apply to new downstream applications. We also provide a visual inspection study to confirm that SpaceVLM retrieves diverse images consistent with negated prompts. We hope the effectiveness of this subspace perspective on VLM embeddings encourages future geometric methods for broader VLM logical reasoning tasks.

\section{Related Work}
\vspace{2pt}\noindent\textbf{Joint Embedding-based Vision-Language Models} align visual and textual representations in a shared embedding space. A representative example is CLIP \cite{radford2021learning}, which trains an image encoder $\mathcal{I}:x\rightarrow \mathbb{R}^d$ and a text encoder $\mathcal{T}:y\rightarrow \mathbb{R}^d$ on $400$ million image-caption pairs using a contrastive objective. The two encoders map inputs to the surface of a unit sphere, and image-text similarity is measured by what we call the \emph{CLIP dot-product scoring} $\mathcal{I}(x) \odot \mathcal{T}(y)$. Given a caption $y$, the corresponding image is then retrieved by: $\hat{x} = \operatorname*{argmax}_x\; \mathcal{I}(x) \odot \mathcal{T}(y)$. 

CLIP’s pretrained encoders are widely used across tasks, from multimodal retrieval \cite{luo2021clip4clip,alpay2023multimodal,lulf2024clip} to multimodal LLMs \cite{liu2023visual,wang2022git} and text-to-image generation \cite{tao2023galip,rombach2022high}. Several follow up variants adopt similar principles: SigLIP \cite{zhai2023sigmoid} replaces the softmax contrastive loss with a sigmoid loss; AIMV2 \cite{fini2025multimodal} replaces the contrastive loss with a multimodal autoregressive loss, but its vision and text encoders can be aligned via Locked-Image Text Tuning \cite{zhai2022lit}, making it applicable to the \emph{CLIP dot-product scoring}. We build on this family of models, improving their handling of negation at inference time without modifying their pretrained parameters.

\vspace{2pt}\noindent\textbf{Fine-tuning for Negation Understanding in VLMs.}
VLMs struggle with logical reasoning in prompts involving conjunction, disjunction, negation, contrast, comparison, condition, causality, and temporality \cite{ma2023crepe, zhou2025logic, li2024genaibench, kang2025clip, li2025exploring}. Most relevant to this work is NegBench \cite{alhamoud2025vision}, which evaluates negation understanding via text-to-image retrieval and image multiple-choice (MCQ) tasks. Most proposed solutions \cite{yuksekgonul2022and,singh2025learning,park2025know} address these problems by constructing logically rich datasets and fine-tuning VLMs on them. NegCLIP \cite{yuksekgonul2022and} fine-tunes CLIP to improve sensitivity to logical structure, including negation. Singh \emph{et al.} \cite{singh2025learning} propose ConCLIP, a CLIP model finetuned for negation understanding, and Alhamoud \emph{et al.} \cite{alhamoud2025vision} extend this research by finetuning CLIP and NegCLIP on CC12M-NegFull, an extension of CC12M \cite{changpinyo2021conceptual} with synthetically augmented negated captions.

While these methods improve negation performance, their reliance on fine-tuning has two drawbacks: (i) degraded zero-shot generalization, and (ii) the fundamental inability of joint embedding–based models to represent negation with a single embedding vector, regardless of the scale of fine-tuning data. Our zero-shot method eliminates both drawbacks by modeling negation geometrically, without any parameter updates.

\vspace{2pt}\noindent\textbf{Towards Training-free Solutions.}
Concurrent to our work, Kang \emph{et al.} \cite{kang2025clip} note that joint embedding–based models cannot geometrically represent negation. They propose DCSM, a modification to the CLIP scoring function that retains all image patch embeddings and text token embeddings, computes cosine similarities across all pairs, and trains a convolutional projection head to aggregate this information. DCSM differs from our approach in two aspects. First, SpaceVLM explicitly models negation as a logical operation through the intersection of subspaces, whereas DCSM does not directly encode logical operators. Second, DCSM trains a lightweight scoring network on top of the frozen CLIP features for each dataset, while our method requires no additional training and operates entirely at inference time.

Few works address negation in text-to-image generation without fine-tuning \cite{lian2024llmgrounded,wu2024self}. They use large language models to parse the negated prompt, construct an intermediate image layout, and feed it to Stable Diffusion with negative prompts to suppress excluded concepts. While effective for generative pipelines, these methods are orthogonal to our goal: they target a specific application and do not improve negation understanding in the underlying vision–language encoder. Other general work \cite{park2025know} also incorporates negation into text–image modeling, but it does so by fine-tuning the CLIP text encoder on specific datasets, which can introduce task-specific specialization and may affect generalization across other negation tasks, as we show in our experiments.

\begin{figure}[tb]
    \begin{subfigure}{0.45\linewidth}
        \centering
        \includegraphics[width=\linewidth]{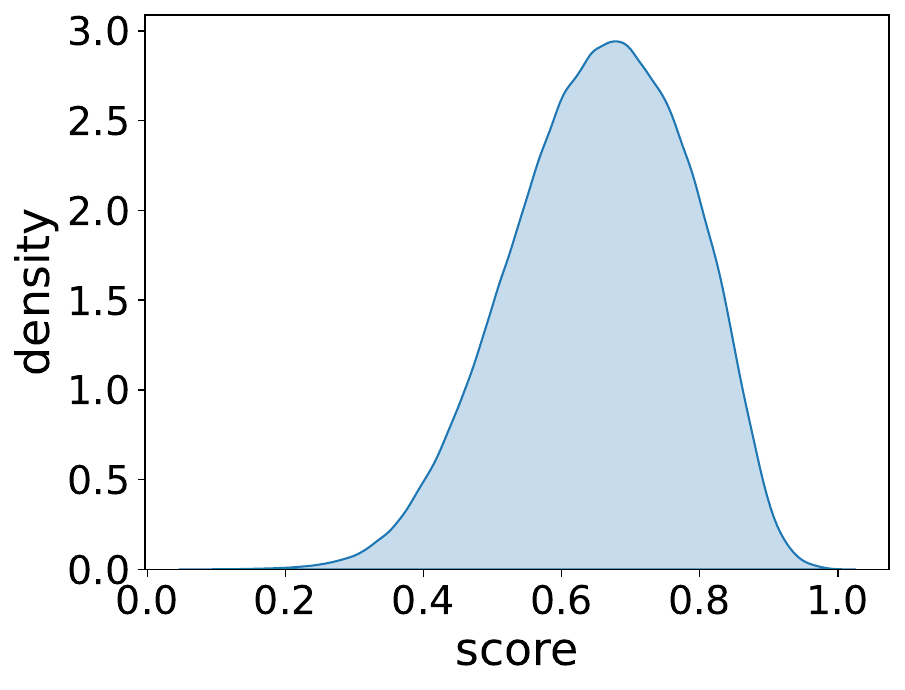}
        \caption{}
        \label{fig1:a}
    \end{subfigure}
    \hfill
    \begin{subfigure}{0.45\linewidth}
        \centering
        \includegraphics[width=\linewidth]{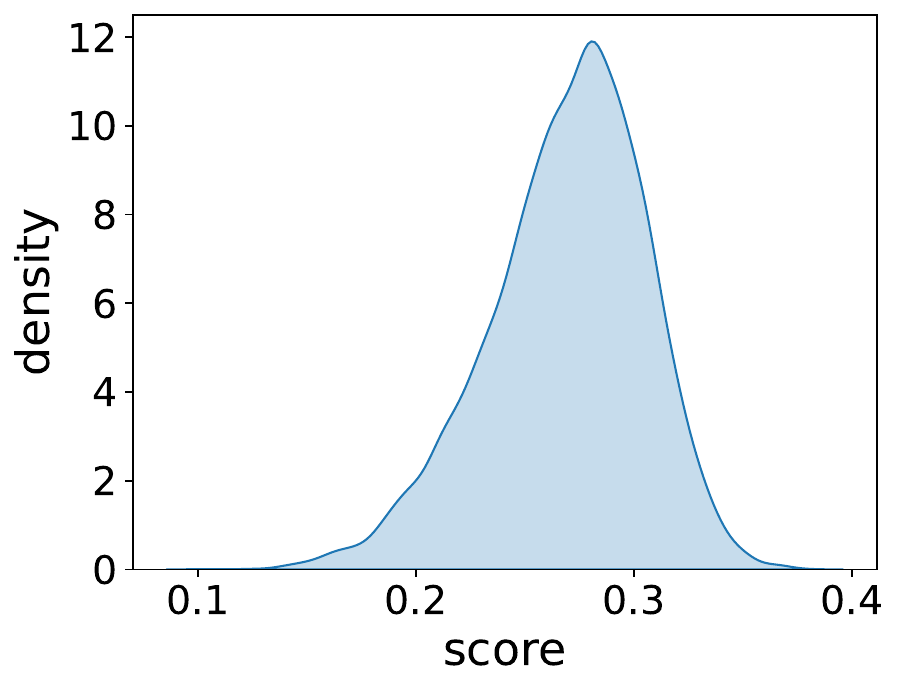}
        \caption{}
        \label{fig1:b}
    \end{subfigure}
    \caption{(a) Cosine similarity score distribution of images within the same category. (b) Cosine similarity score distribution between the textual prompt \texttt{"A photo of a $<$category$>$"} and images belonging to that category.}
    \label{fig:1}
\end{figure}
\section{Method}
We first state the key premise that motivates our approach.

\begin{statement}\label{statement:1}
Negation cannot be modeled by a single point (vector) in the joint embedding VLM space.
\end{statement}

\begin{proof}[Proof sketch]
Let $\mathcal{I}:x\!\to\!\mathbb{R}^d$ and $\mathcal{T}:y\!\to\!\mathbb{R}^d$ be the CLIP image/text encoders with $\|\mathcal{I}(x)\|=\|\mathcal{T}(y)\|=1$. We want to prove that there is no unit vector $n\in\mathbb{R}^d$ that separates cat from non-cat images with a positive margin under the CLIP dot-product scoring; i.e., there do not exist $\beta\in\mathbb{R}$ and $\delta>0$ such that
\[
\inf_{x\ \text{non-cat}}\; n\odot \mathcal{I}(x)\ \ge\ \beta+\delta
\quad\text{and}\quad
\sup_{x\ \text{cat}}\; n\odot \mathcal{I}(x)\ \le\ \beta.
\]

Suppose, for contradiction, that such a unit $n$ and margin $\delta>0$ exist. Then every non-cat image $x$ satisfies $n\odot \mathcal{I}(x)\ge \beta+\delta$. Pick $m$ non-cat images with unit embeddings $u_1,\dots,u_m$ that are pairwise weakly correlated: $u_i\odot u_j\le \gamma$ for $i\neq j$, for some $\gamma\ge 0$ (in high dimension we can choose $\gamma$ arbitrarily small by sampling unrelated classes). Summing the non-cat lower bound gives
\[
m(\beta+\delta)\ \le\ \sum_{i=1}^m n\odot u_i\;=\; n\odot\Big(\sum_{i=1}^m u_i\Big)\ \le\ \Big\|\sum_{i=1}^m u_i\Big\|.
\]
By expanding the norm and using the pairwise bound,
\[
\Big\|\sum_{i=1}^m u_i\Big\|^2
= \sum_{i=1}^m \|u_i\|^2 + 2\!\!\sum_{1\le i<j\le m}\!\! u_i\odot u_j
\ \le\ m + \gamma\, m(m-1),
\]
hence
\[
m(\beta+\delta)\ \le\ \sqrt{\,m + \gamma\, m(m-1)\,}.
\]
Letting $\gamma\to 0$ yields $m(\beta+\delta)\le \sqrt{m}$, i.e. $\beta+\delta \le 1/\sqrt{m}$. As $m\to\infty$, this forces $\beta+\delta\le 0$, contradicting $\delta>0$. Thus, no unit vector $n$ can separate cat from non-cat with any positive margin under the CLIP dot-product scoring. \qedhere
\end{proof}

\subsection{Empirical Divisibility of the Embedding Space}
To model negation more effectively, we first examine the geometric structure of the CLIP embedding space. CLIP aligns images and captions by maximizing cosine similarity, and with $\ell_2$-normalized embeddings, all representations lie on the surface of a $d$-dimensional unit sphere. Empirically, embeddings that refer to the same visual concept (e.g., “dog”) occupy compact, well-separated regions on this sphere \cite{Bhalla2024InterpretingCW, zhao2025quantifying}. 

Figure \ref{fig:1} demonstrates this structure. Figure \ref{fig1:a} shows the distribution of pairwise cosine similarities between images within the same CIFAR-100 class, while Figure \ref{fig1:b} shows similarities between the textual prompt \texttt{"A photo of a <class>"} and images of that class. Both histograms indicate that intra-class samples form high-similarity clusters that are distinct from other concepts. When such clusters are tight and sufficiently separated, we say that the space is \emph{divisible}: a single cosine-similarity threshold can determine whether a new embedding belongs to a concept region or lies outside it. This divisibility property provides the geometric basis for our approach.

\begin{figure}[t]
    \centering
    \includegraphics[width=0.6\linewidth]{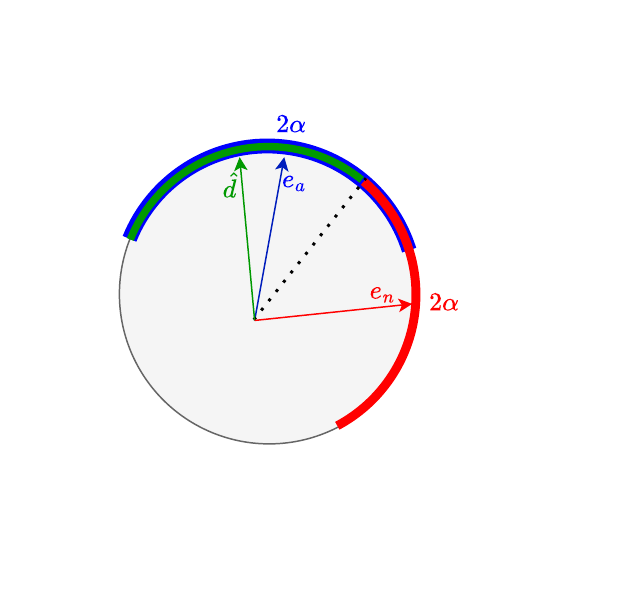}
    \caption{A simple 2D illustration of our approach. Each vector represents the center of its corresponding arc. Given a caption such as \texttt{"A photo of <a> but not <n>"}, $e_a$ denotes the embedding of \texttt{"A photo of <a>"} and $e_n$ denotes the embedding of \texttt{"A photo of <n>"}. We then identify a region that lies close to $e_a$ but distant from $e_n$. The resulting vector $\hat{d}$ serves as the final text embedding, effectively encoding both the affirmative and negated components of the original caption.}
    \label{fig:2}
\end{figure}

\subsection{Problem Formulation}\label{subsec:pf}
Given the divisibility property, we can represent complex textual compositions by reasoning over the regions induced by their constituent concepts.
Consider the input
\[
P = \texttt{"A photo of }{<a>}\texttt{ but not }{<n>}\texttt{"}.
\]
We split $P$ into an affirmative part \(P_a=\texttt{"A photo of }{<a>}\texttt{"}\) and a negated part \(P_n=\texttt{"A photo of }{<n>}\texttt{"}\), so that

\begin{align}\label{eq:split}
P \equiv P_a \;+\; \texttt{"but not"} \;+\; P_n.    
\end{align}

Let
\begin{align}\label{eq:embed}
    e_a=\mathcal{T}(P_a),\qquad  e_n=\mathcal{T}(P_n)
\end{align}
 be the corresponding normalized text embeddings and let \(e_I\) be the normalized image embedding, all produced by CLIP. In standard CLIP inference, the image-text similarity is computed as the dot product \(s=e_I \odot e_P\), where \(e_P\) is the text embedding of the full caption, $P$. However, when \(P\) contains negation, this score becomes unreliable (Statement \ref{statement:1}). Instead, 
our goal is to define a training-free scoring rule that leverages the compact regions (around \(e_a\) and \(e_n\)) to compute a more faithful score.

\begin{table*}[tb]
\centering
\small
\caption{Image/Video retrieval. $\mathrm{R}@K$ uses affirmative queries; $\mathrm{R}@K$–Neg uses negated queries. Our approach narrows (and often closes) the gap between affirmative and negated retrieval when added on top of any baseline.}
\setlength{\tabcolsep}{2.5pt}
\resizebox{\linewidth}{!}{%
\begin{tabular}{cl*{14}{r}}
\toprule
Dataset & Metric
& \multicolumn{2}{c}{CLIP}
& \multicolumn{2}{c}{CLIP\mbox{-}NegFull}
& \multicolumn{2}{c}{ConCLIP}
& \multicolumn{2}{c}{NegCLIP}
& \multicolumn{2}{c}{\makecell[c]{NegCLIP\\\mbox{-}NegFull}}
& \multicolumn{2}{c}{AIMV2}
& \multicolumn{2}{c}{SigLIP-2} \\
\cmidrule(lr){3-4}\cmidrule(lr){5-6}\cmidrule(lr){7-8}\cmidrule(lr){9-10}\cmidrule(lr){11-12}\cmidrule(lr){13-14}\cmidrule(lr){15-16}
& & Base & \cellcolor{gray!20}+ \textbf{Ours}
  & Base & \cellcolor{gray!20}+ \textbf{Ours}
  & Base & \cellcolor{gray!20}+ \textbf{Ours}
  & Base & \cellcolor{gray!20}+ \textbf{Ours}
  & Base & \cellcolor{gray!20}+ \textbf{Ours}
  & Base & \cellcolor{gray!20}+ \textbf{Ours}
  & Base & \cellcolor{gray!20}+ \textbf{Ours}\\
\midrule

\multirow{7}{*}{\rotatebox{90}{COCO}}
& R@1 $\uparrow$
    & \multicolumn{2}{c}{\textcolor{gray}{29.8}}
    & \multicolumn{2}{c}{\textcolor{gray}{32.6}}
    & \multicolumn{2}{c}{\textcolor{gray}{28.8}}
    & \multicolumn{2}{c}{\textcolor{gray}{46.1}}
    & \multicolumn{2}{c}{\textcolor{gray}{44.3}}
    & \multicolumn{2}{c}{\textcolor{gray}{41.0}}
    & \multicolumn{2}{c}{\textcolor{gray}{43.4}}\\
& R@1–Neg $\uparrow$
    & 25.0 & \cellcolor{gray!20}\textbf{29.9}
    & 30.4 & \cellcolor{gray!20}\textbf{33.1}
    & 25.7 & \cellcolor{gray!20}\textbf{28.8}
    & 41.0 & \cellcolor{gray!20}\textbf{45.5}
    & 41.3 & \cellcolor{gray!20}\textbf{44.1}
    & 35.8 & \cellcolor{gray!20}\textbf{41.2}
    & 32.0 & \cellcolor{gray!20}\textbf{43.3}\\
    \cmidrule(lr){2-16}
& R@5 $\uparrow$
    & \multicolumn{2}{c}{\textcolor{gray}{54.8}}
    & \multicolumn{2}{c}{\textcolor{gray}{57.8}}
    & \multicolumn{2}{c}{\textcolor{gray}{54.7}}
    & \multicolumn{2}{c}{\textcolor{gray}{74.0}}
    & \multicolumn{2}{c}{\textcolor{gray}{72.4}}
    & \multicolumn{2}{c}{\textcolor{gray}{66.5}} 
    & \multicolumn{2}{c}{\textcolor{gray}{68.7}}\\
& R@5–Neg $\uparrow$
    & 47.9 & \cellcolor{gray!20}\textbf{55.1}
    & 55.0 & \cellcolor{gray!20}\textbf{58.4}
    & 50.1 & \cellcolor{gray!20}\textbf{54.7}
    & 68.6 & \cellcolor{gray!20}\textbf{73.6}
    & 69.0 & \cellcolor{gray!20}\textbf{71.7}
    & 61.3 & \cellcolor{gray!20}\textbf{66.5}
    & 57.2 & \cellcolor{gray!20}\textbf{68.9}\\
    \cmidrule(lr){2-16}
& R@10 $\uparrow$
    & \multicolumn{2}{c}{\textcolor{gray}{66.0}}
    & \multicolumn{2}{c}{\textcolor{gray}{68.9}}
    & \multicolumn{2}{c}{\textcolor{gray}{66.4}}
    & \multicolumn{2}{c}{\textcolor{gray}{83.0}}
    & \multicolumn{2}{c}{\textcolor{gray}{81.7}}
    & \multicolumn{2}{c}{\textcolor{gray}{76.2}}
    & \multicolumn{2}{c}{\textcolor{gray}{77.8}}\\
& R@10–Neg $\uparrow$
    & 59.1 & \cellcolor{gray!20}\textbf{66.4}
    & 66.5 & \cellcolor{gray!20}\textbf{69.5}
    & 61.8 & \cellcolor{gray!20}\textbf{66.6}
    & 79.0 & \cellcolor{gray!20}\textbf{82.7}
    & 79.1 & \cellcolor{gray!20}\textbf{81.3}
    & 71.9 & \cellcolor{gray!20}\textbf{76.3}
    & 67.9 & \cellcolor{gray!20}\textbf{78.2}\\
    \cmidrule(lr){2-16}
& Avg $\Delta$ (pp)
    & \multicolumn{2}{c}{\cellcolor{gray!20}\textcolor{ForestGreen}{\textbf{+6.5\%}}}
    & \multicolumn{2}{c}{\cellcolor{gray!20}\textcolor{ForestGreen}{\textbf{+3.0\%}}}
    & \multicolumn{2}{c}{\cellcolor{gray!20}\textcolor{ForestGreen}{\textbf{+4.2\%}}}
    & \multicolumn{2}{c}{\cellcolor{gray!20}\textcolor{ForestGreen}{\textbf{+4.4\%}}}
    & \multicolumn{2}{c}{\cellcolor{gray!20}\textcolor{ForestGreen}{\textbf{+2.6\%}}}
    & \multicolumn{2}{c}{\cellcolor{gray!20}\textcolor{ForestGreen}{\textbf{+5.0\%}}}
    & \multicolumn{2}{c}{\cellcolor{gray!20}\textcolor{ForestGreen}{\textbf{+11.1\%}}}\\
\midrule

\multirow{7}{*}{\rotatebox{90}{MSR-VTT}}
& R@1 $\uparrow$
    & \multicolumn{2}{c}{\textcolor{gray}{26.4}}
    & \multicolumn{2}{c}{\textcolor{gray}{28.4}}
    & \multicolumn{2}{c}{\textcolor{gray}{26.4}}
    & \multicolumn{2}{c}{\textcolor{gray}{30.1}}
    & \multicolumn{2}{c}{\textcolor{gray}{30.9}}
    & \multicolumn{2}{c}{\textcolor{gray}{31.1}}
    & \multicolumn{2}{c}{\textcolor{gray}{33.3}}\\
& R@1–Neg $\uparrow$
    & 23.8 & \cellcolor{gray!20}\textbf{26.1}
    & 28.4 & \cellcolor{gray!20}\textbf{28.8}
    & 23.3 & \cellcolor{gray!20}\textbf{25.5}
    & 28.0 & \cellcolor{gray!20}\textbf{30.0}
    & 29.9 & \cellcolor{gray!20}\textbf{29.9}
    & 28.2 & \cellcolor{gray!20}\textbf{30.4}
    & 22.4 & \cellcolor{gray!20}\textbf{27.3}\\
    \cmidrule(lr){2-16}
& R@5 $\uparrow$
    & \multicolumn{2}{c}{\textcolor{gray}{48.7}}
    & \multicolumn{2}{c}{\textcolor{gray}{51.7}}
    & \multicolumn{2}{c}{\textcolor{gray}{48.5}}
    & \multicolumn{2}{c}{\textcolor{gray}{51.5}}
    & \multicolumn{2}{c}{\textcolor{gray}{53.9}}
    & \multicolumn{2}{c}{\textcolor{gray}{52.8}} 
    & \multicolumn{2}{c}{\textcolor{gray}{55.3}} \\
& R@5–Neg $\uparrow$
    & 45.9 & \cellcolor{gray!20}\textbf{49.4}
    & 51.6 & \cellcolor{gray!20}\textbf{52.2}
    & 45.4 & \cellcolor{gray!20}\textbf{50.4}
    & 50.2 & \cellcolor{gray!20}\textbf{52.1}
    & 51.5 & \cellcolor{gray!20}\textbf{53.6}
    & 48.8 & \cellcolor{gray!20}\textbf{52.1}
    & 41.4 & \cellcolor{gray!20}\textbf{48.8}\\
    \cmidrule(lr){2-16}
& R@10 $\uparrow$
    & \multicolumn{2}{c}{\textcolor{gray}{60.7}}
    & \multicolumn{2}{c}{\textcolor{gray}{62.8}}
    & \multicolumn{2}{c}{\textcolor{gray}{61.6}}
    & \multicolumn{2}{c}{\textcolor{gray}{62.1}}
    & \multicolumn{2}{c}{\textcolor{gray}{64.3}}
    & \multicolumn{2}{c}{\textcolor{gray}{62.9}}
    & \multicolumn{2}{c}{\textcolor{gray}{65.7}} \\
& R@10–Neg $\uparrow$
    & 56.6 & \cellcolor{gray!20}\textbf{63.1}
    & 62.9 & \cellcolor{gray!20}\textbf{64.0}
    & 56.4 & \cellcolor{gray!20}\textbf{61.2}
    & 59.7 & \cellcolor{gray!20}\textbf{63.9}
    & 63.7 & \cellcolor{gray!20}\textbf{64.8}
    & 59.4 & \cellcolor{gray!20}\textbf{63.4}
    & 51.6 & \cellcolor{gray!20}\textbf{60.8}\\
    \cmidrule(lr){2-16}
& Avg $\Delta$ (pp)
    & \multicolumn{2}{c}{\cellcolor{gray!20}\textcolor{ForestGreen}{\textbf{+4.1\%}}}
    & \multicolumn{2}{c}{\cellcolor{gray!20}\textcolor{ForestGreen}{\textbf{+0.7\%}}}
    & \multicolumn{2}{c}{\cellcolor{gray!20}\textcolor{ForestGreen}{\textbf{+4.0\%}}}
    & \multicolumn{2}{c}{\cellcolor{gray!20}\textcolor{ForestGreen}{\textbf{+2.7\%}}}
    & \multicolumn{2}{c}{\cellcolor{gray!20}\textcolor{ForestGreen}{\textbf{+1.1\%}}}
    & \multicolumn{2}{c}{\cellcolor{gray!20}\textcolor{ForestGreen}{\textbf{+3.2\%}}}
     & \multicolumn{2}{c}{\cellcolor{gray!20}\textcolor{ForestGreen}{\textbf{+7.2\%}}}\\
\bottomrule
\end{tabular}}
\label{tab:retrieval}
\end{table*}

\begin{figure}[t]
\captionsetup{labelformat=empty}  
\caption{\textbf{Algorithm 1:} PyTorch-style pseudocode for SpaceVLM, which computes negation-aware text embeddings for a generic VLM.
}
\lstset{style=pytorch}
\begin{lstlisting}
# Inputs:
#   caption, text_encoder, LLM, threshold t in [-1, 1]
# Output:
#   d_hat : negation-aware embedding of input caption

# 1. Split into affirmative and negative  (Eq. 1)
aff_cap, neg_cap = LLM(caption)
    
# 2. Encode using the original VLM encoder  (Eq. 2)
e_a = text_encoder(aff_cap)
e_b = text_encoder(neg_cap)
    
# 3. Compute angular distances  (Eq. 4)
alpha = arccos(threshold)
theta = arccos(dot_product(aff_embed, neg_embed))
    
# 4. Compute negtion-aware embedding  (Eq. 3)
d_hat = aff_embed * sin(alpha + theta/2) / sin(theta)
d_hat -= neg_embed * sin(alpha - theta/2) / sin(theta) 
    
# 5. Normalize  (Eq. 5)
d_hat = d_hat / norm(d_hat)
\end{lstlisting}
\label{alg:1}
\end{figure}

\subsection{SpaceVLM: Sub-Space Modeling of Negation}
We now define the training-free scoring rule that models negation as a subspace. We start with the affirmative and negated embeddings $e_a=\mathcal{T}(P_a)$ and $e_n=\mathcal{T}(P_n)$. In practice, a language processor such as a lightweight LLM is used to split input text into its affirmative and negative parts $P_a$ and $P_n$. Note that both $P_a$ and $P_n$ are phrased as affirmative captions. We denote the neighborhood (spherical cap) of a normalized point $x$ in the VLM space as
\[
    \mathcal{N}(x)=\{\,z\in\mathbb{R}^d \mid x\odot z \ge t\,\}, \qquad t\in[-1,1],
\]
where $t$ is a cosine-similarity threshold. We associate $P_a$ and $P_n$ with their subspaces $\mathcal{N}(e_a)$ and $\mathcal{N}(e_n)$. The target subspace for $P$ is the region that is close to the affirmative concept but outside the neighborhood of the negated one:
\[
    \mathcal{N}(P) \;=\; \mathcal{N}(e_a)\,\cap\,\mathcal{N}^{c}(e_n),
\]
where
\[
    \mathcal{N}^{c}(e_n) \;=\; \{\,z\in\mathbb{R}^d \mid z\notin \mathcal{N}(e_n)\,\}.
\]

To perform image-text matching, we need a similarity score between an image embedding $e_I$ and this region $\mathcal{N}(P)$. While one could measure distances from a point to a (curved) region, CLIP’s geometry suggests a simpler surrogate: because embeddings lie on a unit sphere and cosine similarity is rotationally symmetric, a representative \emph{direction} for $\mathcal{N}(P)$ provides a natural scoring vector. We choose the direction $\hat{d}$ at the angular “center” of the feasible region:
\begin{align}\label{eq:d_hat}
    \hat d
    = \frac{\sin(\alpha + \frac{\theta}{2})}{\sin(\theta)}\, e_a +
    \frac{\sin(\alpha - \frac{\theta}{2})}{\sin(\theta)}\, e_n
\end{align}
where
\begin{align}\label{eq:hyper}
    \alpha=\arccos(t),\qquad
\theta=\arccos(e_a\odot e_n).
\end{align}

Intuitively, $\theta$ is the angle between $e_a$ and $e_n$, and $\alpha$ defines the cap radius induced by the threshold $t$. The vector $\hat d$ points to the center of the intersection region $\mathcal{N}(e_a)\cap\mathcal{N}^c(e_n)$ along the great-circle arc joining $e_n$ and $e_a$.

The final score uses the standard CLIP dot-product form with this direction (optionally normalized):
\begin{align}\label{eq:d_hat_norm}
    \tilde d \;=\; \frac{\hat d}{\|\hat d\|}, \qquad
s_{\text{neg}}(e_I,P) \;=\; e_I \odot \tilde d.
\end{align}

Algorithm \ref{alg:1} provides the pseudocode of our subspace method for computing negation-aware embeddings.

\section{Experiments}
We evaluate the effectiveness of our approach for enhancing negation understanding across multiple VLMs.

\subsection{Evaluation Protocol}
\vspace{2pt}\noindent\textbf{Tasks.}
Following the \emph{NegBench} \cite{alhamoud2025vision}, we assess negation understanding on two tasks:
\begin{inparaenum}[(i)]
    \item \textbf{Image/Video Retrieval} with \emph{negated} queries, and
    \item \textbf{Text Retrieval (MCQ)} with \emph{negated} captions.
\end{inparaenum}
The negated retrieval task measures coarse-grained reasoning: given a negated query such as, \texttt{"A photo of a dog not on grass,"} the model must retrieve relevant images or videos (\Cref{fig:task_overview}a). The MCQ task measures fine-grained reasoning: given an image, the model selects the correct caption among four closely related candidates drawn from \emph{Affirmation}, \emph{Negation}, and \emph{Hybrid} templates. An example is visualized in \Cref{fig:task_overview}c.

For medical VLMs, NegBench includes a simplified MCQ task providing a binary choice between negated and affirmative captions: for instance, \texttt{"This image shows Lung Opacity"} vs. \texttt{"This image does not show Lung Opacity."} We later extend this evaluation to a new text-to-image generation (T2I) task.

\vspace{2pt}\noindent\textbf{Datasets.}
For negated Image/Video Retrieval, we use the negated extensions of COCO \cite{lin2014microsoft} and MSR-VTT \cite{xu2016msr} provided by NegBench \cite{alhamoud2025vision}.  For MCQ, samples are drawn from COCO, VOC-2007 \cite{pascal-voc-2007}, and MSR-VTT. For MCQ in the medical domain, we use negated CheXpert \cite{irvin2019chexpert, alhamoud2025vision}.

\begin{table*}[t]
\centering
\small
\caption{MCQ results showing the effect of our approach on each VLM across Affirmative, Negation, and Hybrid templates. Our method achieves an average improvement of over 30\% across all datasets.}
\setlength{\tabcolsep}{2.5pt} 
\resizebox{\linewidth}{!}{%
\begin{tabular}{cl*{14}{r}}
\toprule
Dataset & Metric $\uparrow$
& \multicolumn{2}{c}{CLIP}
& \multicolumn{2}{c}{CLIP\mbox{-}NegFull}
& \multicolumn{2}{c}{ConCLIP}
& \multicolumn{2}{c}{NegCLIP}
& \multicolumn{2}{c}{NegCLIP\mbox{-}NegFull}
& \multicolumn{2}{c}{AIMV2}
& \multicolumn{2}{c}{SigLIP-2} \\
\cmidrule(lr){3-4}\cmidrule(lr){5-6}\cmidrule(lr){7-8}\cmidrule(lr){9-10}\cmidrule(lr){11-12}\cmidrule(lr){13-14}\cmidrule(lr){15-16}
& & Base & \cellcolor{gray!20}+ \textbf{Ours} & Base & \cellcolor{gray!20}+ \textbf{Ours}
  & Base & \cellcolor{gray!20}+ \textbf{Ours} & Base & \cellcolor{gray!20}+ \textbf{Ours}
  & Base & \cellcolor{gray!20}+ \textbf{Ours} & Base & \cellcolor{gray!20}+ \textbf{Ours}  & Base & \cellcolor{gray!20}+ \textbf{Ours} \\
\midrule

\multirow{4}{*}{\rotatebox{90}{COCO}}
& Affirmative & 70.0 & \cellcolor{gray!20}\textbf{77.4} & 73.1 & \cellcolor{gray!20}\textbf{83.8} & 15.6 & \cellcolor{gray!20}\textbf{52.0} & 49.2 & \cellcolor{gray!20}\textbf{63.5} & 81.0 & \cellcolor{gray!20}\textbf{78.5} & 52.1 & \cellcolor{gray!20}\textbf{78.6} & 45.8 & \cellcolor{gray!20}\textbf{75.1} \\
& Negation   &  6.6 & \cellcolor{gray!20}\textbf{71.8} & 33.2 & \cellcolor{gray!20}\textbf{69.3} & 32.9 & \cellcolor{gray!20}\textbf{78.2} & 13.9 & \cellcolor{gray!20}\textbf{78.4} & 25.9 & \cellcolor{gray!20}\textbf{74.5} & 16.6 & \cellcolor{gray!20}\textbf{71.3} & 9.2 & \cellcolor{gray!20}\textbf{70.3}\\
& Hybrid     & 38.4 & \cellcolor{gray!20}\textbf{50.0} & 54.7 & \cellcolor{gray!20}\textbf{55.5} & 25.3 & \cellcolor{gray!20}\textbf{45.8} & 16.3 & \cellcolor{gray!20}\textbf{51.7} & 60.1 & \cellcolor{gray!20}\textbf{53.6} & 30.5 & \cellcolor{gray!20}\textbf{49.5} & 32.8 & \cellcolor{gray!20}\textbf{45.7} \\
& \textbf{AVG}        & 39.2 & \cellcolor{gray!20}{\textcolor{ForestGreen}{\textbf{\scriptsize(+27.1\%)}}\,\textbf{66.3}}
               & 54.2 & \cellcolor{gray!20}{\textcolor{ForestGreen}{\textbf{\scriptsize(+15.4\%)}}\,\textbf{69.6}}
               & 24.4 & \cellcolor{gray!20}{\textcolor{ForestGreen}{\textbf{\scriptsize(+28.8\%)}}\,\textbf{58.2}}
               & 26.8 & \cellcolor{gray!20}{\textcolor{ForestGreen}{\textbf{\scriptsize(+37.4\%)}}\,\textbf{64.2}}
               & 56.5 & \cellcolor{gray!20}{\textcolor{ForestGreen}{\textbf{\scriptsize(+12.3\%)}}\,\textbf{68.8}}
               & 33.5 & \cellcolor{gray!20}{\textcolor{ForestGreen}{\textbf{\scriptsize(+32.9\%)}}\,\textbf{66.4}}
               & 29.8 & \cellcolor{gray!20}{\textcolor{ForestGreen}{\textbf{\scriptsize(+33.8\%)}}\,\textbf{63.6}} \\
\midrule

\multirow{4}{*}{\rotatebox{90}{VOC2007}}
& Affirmative & 80.9 & \cellcolor{gray!20}\textbf{85.8} & 85.0 & \cellcolor{gray!20}\textbf{91.2} & 24.8 & \cellcolor{gray!20}\textbf{66.1} & 70.5 & \cellcolor{gray!20}\textbf{80.1} & 81.0 & \cellcolor{gray!20}\textbf{87.8} & 64.5 & \cellcolor{gray!20}\textbf{89.1} & 55.9 & \cellcolor{gray!20}\textbf{85.5} \\
& Negation   &  3.0 & \cellcolor{gray!20}\textbf{84.2} & 31.7 & \cellcolor{gray!20}\textbf{81.0} & 23.2 & \cellcolor{gray!20}\textbf{83.9} &  4.6 & \cellcolor{gray!20}\textbf{90.1} & 21.1 & \cellcolor{gray!20}\textbf{80.7} & 9.1 & \cellcolor{gray!20}\textbf{77.7} & 3.8 & \cellcolor{gray!20}\textbf{75.6} \\
& Hybrid     & 58.0 & \cellcolor{gray!20}\textbf{76.8} & 79.5 & \cellcolor{gray!20}\textbf{80.8} & 56.7 & \cellcolor{gray!20}\textbf{79.4} & 42.3 & \cellcolor{gray!20}\textbf{82.4} & 83.7 & \cellcolor{gray!20}\textbf{88.1} & 42.6 & \cellcolor{gray!20}\textbf{71.2} & 39.9 & \cellcolor{gray!20}\textbf{68.3} \\
& \textbf{AVG}        & 37.9 & \cellcolor{gray!20}{\textcolor{ForestGreen}{\textbf{\scriptsize(+44.8\%)}}\,\textbf{81.1}}
               & 60.1 & \cellcolor{gray!20}{\textcolor{ForestGreen}{\textbf{\scriptsize(+22.2\%)}}\,\textbf{82.3}}
               & 38.2 & \cellcolor{gray!20}{\textcolor{ForestGreen}{\textbf{\scriptsize(+41.3\%)}}\,\textbf{79.5}}
               & 30.2 & \cellcolor{gray!20}{\textcolor{ForestGreen}{\textbf{\scriptsize(+55.1\%)}}\,\textbf{85.3}}
               & 58.2 & \cellcolor{gray!20}{\textcolor{ForestGreen}{\textbf{\scriptsize(+34.5\%)}}\,\textbf{84.7}}
               & 31.4 & \cellcolor{gray!20}{\textcolor{ForestGreen}{\textbf{\scriptsize(+44.9\%)}}\,\textbf{76.3}} 
               & 26.8 & \cellcolor{gray!20}{\textcolor{ForestGreen}{\textbf{\scriptsize(+46.9\%)}}\,\textbf{73.7}} \\
\midrule

\multirow{4}{*}{\rotatebox{90}{MSR-VTT}}
& Affirmative & 60.9 & \cellcolor{gray!20}\textbf{81.2} & 81.8 & \cellcolor{gray!20}\textbf{87.2} & 57.3 & \cellcolor{gray!20}\textbf{75.8} & 47.5 & \cellcolor{gray!20}\textbf{75.5} & 75.8 & \cellcolor{gray!20}\textbf{85.4} & 57.3 & \cellcolor{gray!20}\textbf{78.8} & 52.2 & \cellcolor{gray!20}\textbf{79.1} \\
& Negation   & 15.6 & \cellcolor{gray!20}\textbf{34.0} & 21.2 & \cellcolor{gray!20}\textbf{27.8} & 30.3 & \cellcolor{gray!20}\textbf{44.2} & 13.6 & \cellcolor{gray!20}\textbf{51.6} & 22.1 & \cellcolor{gray!20}\textbf{41.1} & 6.0 & \cellcolor{gray!20}\textbf{41.9} & 8.8 & \cellcolor{gray!20}\textbf{41.9} \\
& Hybrid     & 18.3 & \cellcolor{gray!20}\textbf{60.3} & 35.6 & \cellcolor{gray!20}\textbf{66.0} & 38.8 & \cellcolor{gray!20}\textbf{68.3} & 16.3 & \cellcolor{gray!20}\textbf{60.9} & 35.3 & \cellcolor{gray!20}\textbf{66.3} & 22.8 & \cellcolor{gray!20}\textbf{59.2} & 21.2 & \cellcolor{gray!20}\textbf{57.7} \\
& \textbf{AVG}        & 31.6 & \cellcolor{gray!20}{\textcolor{ForestGreen}{\textbf{\scriptsize(+27.6\%)}}\,\textbf{58.0}}
               & 46.0 & \cellcolor{gray!20}{\textcolor{ForestGreen}{\textbf{\scriptsize(+13.6\%)}}\,\textbf{59.6}}
               & 42.0 & \cellcolor{gray!20}{\textcolor{ForestGreen}{\textbf{\scriptsize(+20.3\%)}}\,\textbf{62.3}}
               & 25.8 & \cellcolor{gray!20}{\textcolor{ForestGreen}{\textbf{\scriptsize(+36.7\%)}}\,\textbf{62.5}}
               & 44.2 & \cellcolor{gray!20}{\textcolor{ForestGreen}{\textbf{\scriptsize(+19.6\%)}}\,\textbf{63.8}}
               & 28.4 & \cellcolor{gray!20}{\textcolor{ForestGreen}{\textbf{\scriptsize(+30.8\%)}}\,\textbf{59.2}}
               & 27.2 & \cellcolor{gray!20}{\textcolor{ForestGreen}{\textbf{\scriptsize(+32.1\%)}}\,\textbf{59.3}}\\
\bottomrule
\end{tabular}}
\label{tab:mcq}
\end{table*}

\vspace{2pt}\noindent\textbf{Metrics.}
For retrieval, we report Recall@K ($\mathrm{R}@K$ for $K\in\{1,5,10\}$), measuring the fraction of queries where at least one relevant image or video appears in the top-K results. We report performance for both standard (affirmative) and negated queries. 
For MCQ, we report accuracy, decomposed by the template of the correct answer (\emph{Affirmation}, \emph{Negation}, \emph{Hybrid}), to expose performance gaps between affirmative and negated captions. For binary MCQs on medical VLMs, we report accuracy only, since there are only two possible options for each image. For text-to-image generation, accuracy measures whether the generated image successfully excludes the object negated in the input prompt.

\vspace{2pt}\noindent\textbf{Hyperparameters.}
The similarity threshold $t$ is tuned per dataset on validation splits. As shown in Subsection~\ref{subsec:as}, optimal $t$ values lie in $[0.90, 0.95]$ across all datasets, and performance is robust within this range (negligible degradation), enabling simple hand-set choices of $t$ in new downstream applications without expensive tuning. To decompose input queries into affirmative and negated parts, we use a lightweight language processor based on Mistral-7B-v0.3 \cite{jiang2023mistral7b}, fine-tuned on small subsets of COCO (Image Neg-Retrieval) and VOC-2007 (MCQ) \cite{alhamoud2025vision}. This module \emph{does not} modify the VLM and is used solely for query decomposition. In Subsection \ref{subsec:as}, we compare different LLMs with different sizes with respect to final performance and inference time.

\vspace{2pt}\noindent\textbf{Baselines.}
We evaluate our method added to nine models spanning both pretrained and fine-tuned VLMs. Pretrained VLMs include CLIP \cite{radford2021learning}, AIMV2 \cite{fini2025multimodal}, and SigLIP-2 \cite{tschannen2025siglip}. Fine-tuned variants include \textbf{CLIP-NegFull} (fine-tuned on CC12M-NegFull \cite{alhamoud2025vision}), \textbf{ConCLIP} (fine-tuned on CC-Neg \cite{singh2025learning}), \textbf{NegCLIP} (fine-tuned on COCO with hard negative captions), and \textbf{NegCLIP-NegFull} (fine-tuned on CC12M-NegFull). For the medical MCQ task, we use \textbf{BiomedCLIP} \cite{zhang2023biomedclip}. We apply our training-free and model-agnostic method directly to each baseline and report results with and without our modification to isolate its effect. Unless otherwise stated, all models use the ViT-B/32 backbone for consistency. 

\subsection{Evaluation on NegBench}
\vspace{2pt}\noindent\textbf{Image/Video Retrieval.}
Table~\ref{tab:retrieval} reports results for standard retrieval ($\mathrm{R}@K$; non-negated queries) and negated retrieval ($\mathrm{R}@K$–Neg; negated queries). The $\mathrm{R}@K$ accuracy of each model serves as an approximate upper bound for its negated counterpart $\mathrm{R}@K$–Neg. 

SpaceVLM improves retrieval across all baselines and datasets, substantially closing the gap between affirmative and negated queries. In some cases, retrieval performance on negated queries even exceeds that of the base model on standard queries, as the additional negation information helps disambiguate similar images and pick the most accurate one. Importantly, performance on affirmative queries remains unchanged, confirming that our scoring rule preserves the original model behavior on non-negated prompts.

\vspace{2pt}\noindent\textbf{MCQ.}
Table~\ref{tab:mcq} shows MCQ results. Across all models and datasets, SpaceVLM has large gains, especially when the correct caption follows a \emph{Negation} template. More surprisingly, it also improves accuracy when the correct caption follows an \emph{Affirmation} template. This is because it reduces confusion with other templates. For example, vanilla CLIP maps both captions \texttt{"a photo of a fish and coral"} and \texttt{"a photo of a fish but not coral"} to nearly identical embeddings, which causes the model to select them interchangeably (\Cref{fig:task_overview}c). With our geometric scoring, these captions become clearly separable. Notably, when applied to vanilla CLIP, our method outperforms several fine-tuned baselines trained specifically for negation understanding, such as CLIP-NegFull. Since MCQ is the most fine-grained and challenging diagnostic test in NegBench \cite{alhamoud2025vision}, this result demonstrates that our approach resolves the core failure mode of VLMs without fine-tuning.

\vspace{2pt}\noindent\textbf{Binary MCQs and medical VLMs.} We apply our method to improve the accuracy of BiomedCLIP given medical negations in the CheXpert MCQ task \cite{alhamoud2025vision}. The control task includes affirmative captions only, whereas the negated dataset includes both an affirmative and a negated caption for each image. The results, shown in Table 3, suggest that our training-free SpaceVLM readily generalizes to specialized domains, such as healthcare.

\begin{center}
\vspace{-0.5em}
\tablestyle{2pt}{1.2}
\begin{tabular}{c|x{80}x{80}}
Model & CheXpert-Control & CheXpert-Negation \\
\shline
BiomedCLIP & 66.8 & 45.5 \\
\quad$\rightarrow$ + Ours & 66.8 & 67.4 (\emphhigh{$\uparrow$21.9\%}) \\
\end{tabular}
\end{center}

\vspace{2pt}\noindent\textbf{Comparison with Concurrent Works.}
DCSM \cite{kang2025clip} is a concurrent method that also targets the inability of joint embedding-based VLMs, such as CLIP, to represent negation using a single vector. An important step in DCSM is learning a projection layer specific to each model–dataset pair, requiring additional training for every new domain. Another concurrent work is NegationCLIP \cite{park2025know}, which introduces a new benchmark for negation and fine-tunes CLIP on it. Their benchmark is not yet available, and they do not evaluate on NegBench. 

We evaluate NegationCLIP on NegBench MCQ tasks, along with reported DCSM results on COCO and VOC-2007 MCQ tasks. Using the same CLIP ViT-B/16 backbone across all models, our SpaceVLM scoring achieves substantially higher accuracy on both datasets, with no training required. 

\begin{center}
\vspace{-0.5em}
\tablestyle{2pt}{1.2}
\begin{tabular}{c|x{58}x{58}}
Method & COCO MCQ & VOC2007 MCQ \\
\shline
DCSM \cite{kang2025clip} & 48.6 & 49.0 \\
NegationCLIP \cite{park2025know} & 29.8 & 38.8 \\
\textbf{SpaceVLM (ours)} & \textbf{68.1}  & \textbf{78.5} \\
\end{tabular}
\end{center}

\subsection{Application to Text-to-Image Generation (T2I)}
\vspace{2pt}\noindent\textbf{Experimental Setup.}
We test whether the proposed SpaceVLM scoring improves negation adherence in T2I generation systems. CLIP text encoders are widely used in modern T2I models \cite{rombach2022high,tao2023galip}, yet their limited handling of negation often causes generated images to exhibit objects explicitly excluded in the prompt. We apply our method to GALIP \cite{tao2023galip}, a GAN-based generator that uses a CLIP text encoder and produces image quality comparable to Stable Diffusion \cite{rombach2022high} and Matching Flows \cite{lipman2022flow}. We evaluate on the 107 negated prompts from \cite{park2025know}, which cover diverse negation types, and use Gemma-3-27B-it \cite{team2025gemma} as an automatic evaluator for presence/absence checks. We focus on GALIP for clarity of analysis; diffusion models such as Stable Diffusion condition on token-level embeddings, while our method produces a single text embedding. Extending SpaceVLM to token-wise conditioning is left for future work.

Table \ref{tab:t2i} reports the T2I results. Our method substantially improves negation adherence, with up to 37\% higher accuracy over baselines. Intuitively, these gains arise because the subspace formulation explicitly removes the negated concept from the CLIP text embedding, enabling the generator to condition on representations that better match the intended prompt semantics (\Cref{fig:task_overview}b).

\begin{table}[tb]
\centering
\small
\renewcommand{\arraystretch}{0.9}
\caption{Text-to-Image generation. \emph{Affirmative Acc} is the accuracy of correctly generating the positive concept; \emph{Negation Acc} is the accuracy of omitting the negated concept; \emph{Acc} requires both to be satisfied simultaneously.}

\begin{tabular}{llll}
\toprule

Model & Aff-Acc $\uparrow$ & Neg-Acc $\uparrow$ & Acc $\uparrow$\\
\midrule
  
  CLIP & 97.3 & 28.5 & 27.3\\
  \cellcolor{gray!20}\quad$\rightarrow$ + Ours & \cellcolor{gray!20}98.8 & \cellcolor{gray!20}60.9 & \cellcolor{gray!20}59.7 \textcolor{ForestGreen}{\textbf{\scriptsize(+32.4 $\uparrow$)}} \\
  \addlinespace[1.0ex]
  
  CLIP-NegFull & 98.1 & 40.7 & 39.7\\
  \cellcolor{gray!20}\quad$\rightarrow$ + Ours & \cellcolor{gray!20}97.4 & \cellcolor{gray!20}\cellcolor{gray!20}64.0 & \cellcolor{gray!20}61.8 \textcolor{ForestGreen}{\textbf{\scriptsize(+22.1 $\uparrow$)}}\\
  \addlinespace[1.0ex]
  
  ConCLIP & 27.7 & 68.3 & 11.0\\
  \cellcolor{gray!20}\quad$\rightarrow$ + Ours & \cellcolor{gray!20}86.6 &\cellcolor{gray!20}57.8 & \cellcolor{gray!20}48.2 \textcolor{ForestGreen}{\textbf{\scriptsize(+37.2 $\uparrow$)}} \\
  \addlinespace[1.0ex]
  
  NegCLIP & 98.8 & 24.5 & 23.7 \\
  \cellcolor{gray!20}\quad$\rightarrow$ + Ours  & \cellcolor{gray!20}98.9 & \cellcolor{gray!20}60.6 & \cellcolor{gray!20}59.8 \textcolor{ForestGreen}{\textbf{\scriptsize(+36.1 $\uparrow$)}} \\
  \addlinespace[1.0ex]
  
  NegCLIP-NegFull & 98.6 & 35.5 & 34.8 \\
  \cellcolor{gray!20}\quad$\rightarrow$ + Ours & \cellcolor{gray!20}98.0 & \cellcolor{gray!20}63.9 & \cellcolor{gray!20}62.3 \textcolor{ForestGreen}{\textbf{\scriptsize(+27.5 $\uparrow$)}}\\
  \addlinespace[1.0ex]
  
  NegationCLIP & 98.8 & 45.2 & 44.5 \\
  \cellcolor{gray!20}\quad$\rightarrow$ + Ours & \cellcolor{gray!20}97.1 & \cellcolor{gray!20}60.7 & \cellcolor{gray!20}58.6 \textcolor{ForestGreen}{\textbf{\scriptsize(+14.1 $\uparrow$)}}\\
  
\bottomrule
\end{tabular}
\label{tab:t2i}
\end{table}

\vspace{-4pt}
\subsection{Ablation Studies}\label{subsec:as}
\begin{figure}
    \centering
    \includegraphics[width=\linewidth, height=4cm]{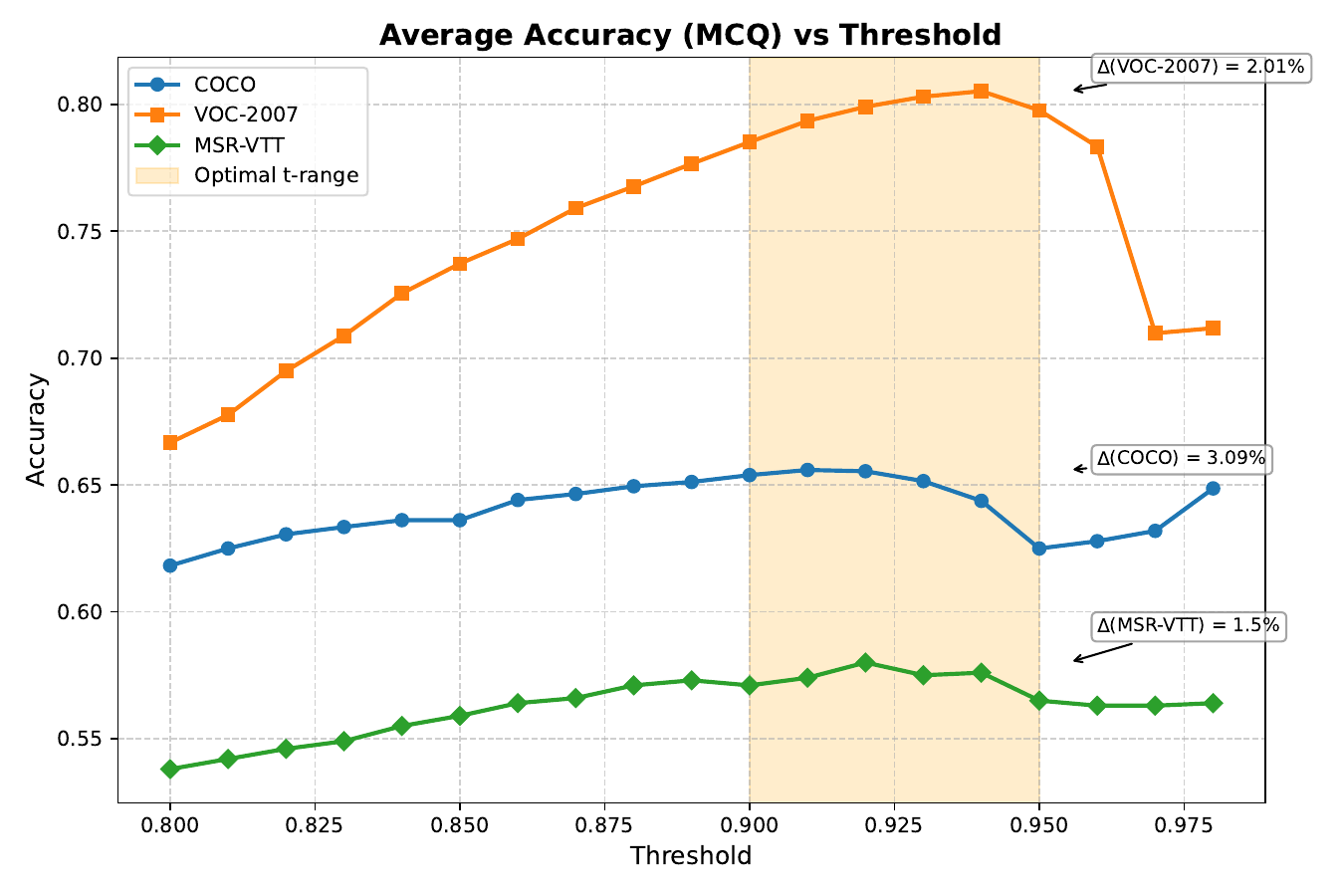}
    \caption{Effect of varying threshold $t$ on MCQ average accuracy. The range $[0.90, 0.95]$ is near-optimal with at most a $3.09\%$ drop on COCO. For best results, perform cross-validation on the target task.}
    
    \label{fig:th}
\end{figure}

\vspace{2pt}\noindent\textbf{Varying VLM Complexity}
We evaluate how VLM complexity affects our method by testing three backbones of increasing size: ViT-B/32, ViT-B/16, and ViT-L/14. Using the MCQ COCO task, the inline table below reports average results. Across all model sizes, our method consistently improves performance, indicating its robustness and applicability to VLMs of varying capacity.
\begin{center}
\vspace{-0.5em}
\tablestyle{2pt}{1.2}
\resizebox{\linewidth}{!}{
\begin{tabular}{c|x{60}x{60}x{60}}
Model & ViT-B/32 & ViT-B/16 & ViT-L/14 \\
\shline
CLIP & 39.2 & 41.4 & 38.5 \\
\quad$\rightarrow$ + \textbf{Ours} & \textbf{66.3} (\emphhigh{$\uparrow$27.1\%}) & \textbf{67.4} (\emphhigh{$\uparrow$26.0\%}) & \textbf{65.9} (\emphhigh{$\uparrow$27.4\%}) \\
\end{tabular}}
\end{center}
\vspace{2pt}\noindent\textbf{Sensitivity to the Threshold $t$.}
We analyze performance sensitivity to $t$ on the MCQ benchmark by varying $t\in[0.90,0.95]$ (Figure~\ref{fig:th}). The maximum drop is $3.09\%$ (COCO), indicating robustness and enabling practical, hand-set choices of $t$ in many applications. For the highest accuracy, we recommend cross-validation on the target task.

\begin{figure}
    \centering
    \includegraphics[width=\linewidth]{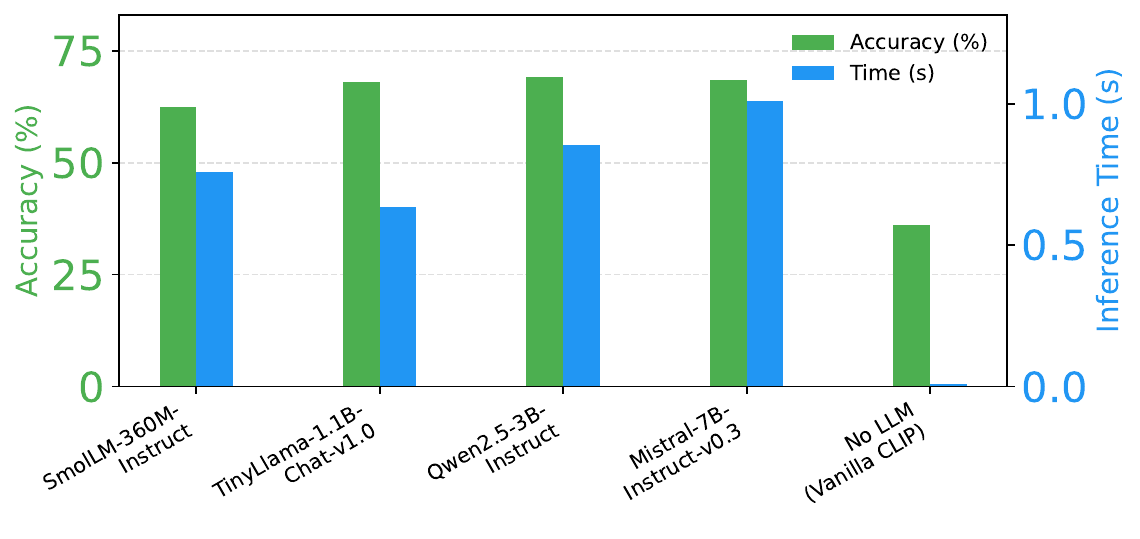}
    \caption{Accuracy vs Inference time tradeoff among different LLMs as a pre-processor for our method.}
    \label{fig:llm}
\end{figure}

\vspace{2pt}\noindent\textbf{Language Pre-processor}
We study how the choice of LLM, used to split captions into affirmative and negative parts, affects computation time and downstream performance. We ablate across several LLMs of varying scale and capability, evaluating SmolLM-360M-Instruct \cite{allal2024SmolLM}, TinyLlama-1.1B-Chat-v1.0 \cite{zhang2024TinyLlama}, Qwen2.5-3B-Instruct \cite{qwen2.5}, and Mistral \cite{mistral7binstruct_v0.3}.

Figure \ref{fig:llm} shows this comparison. We evaluate performance on the NegBench MCQ tasks by reporting average accuracy of SpaceVLM across COCO, VOC2007, and MSR-VTT. For inference time (the time required to obtain the text embedding for a given caption), we similarly average across datasets. The inference time is computed for a 32-input batch on a single H100 GPU, which is a standard in cloud computing. TinyLlama-1B provides a favorable balance between accuracy and inference speed relative to the other models, which makes it a practical choice in real-world settings.

\vspace{-11pt}
\subsection{Visualization}
\begin{figure}[tb]
    \centering
    \begin{subfigure}{0.8\linewidth}
        \centering
        \includegraphics[width=\linewidth, height=3cm]{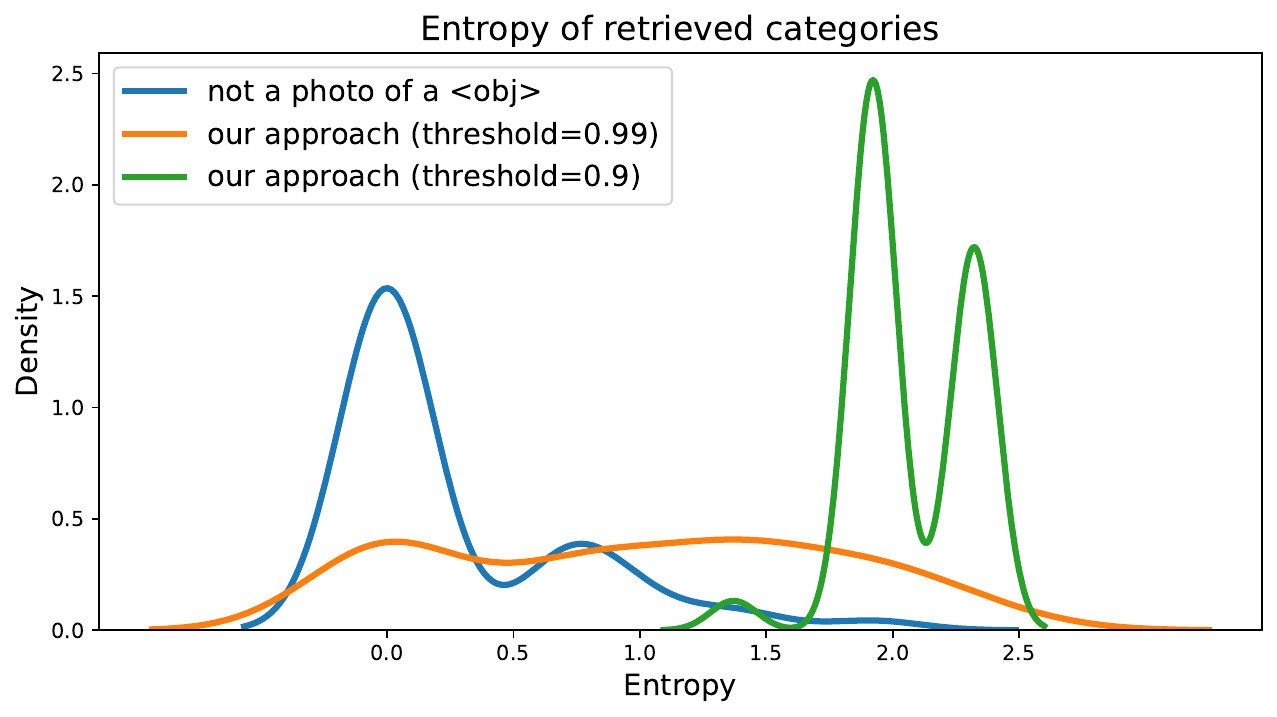}
        \caption{}
        \label{fig3:a}
    \end{subfigure}
        
    \vfill

    \begin{subfigure}{0.8\linewidth}
        \centering
        \includegraphics[width=\linewidth]{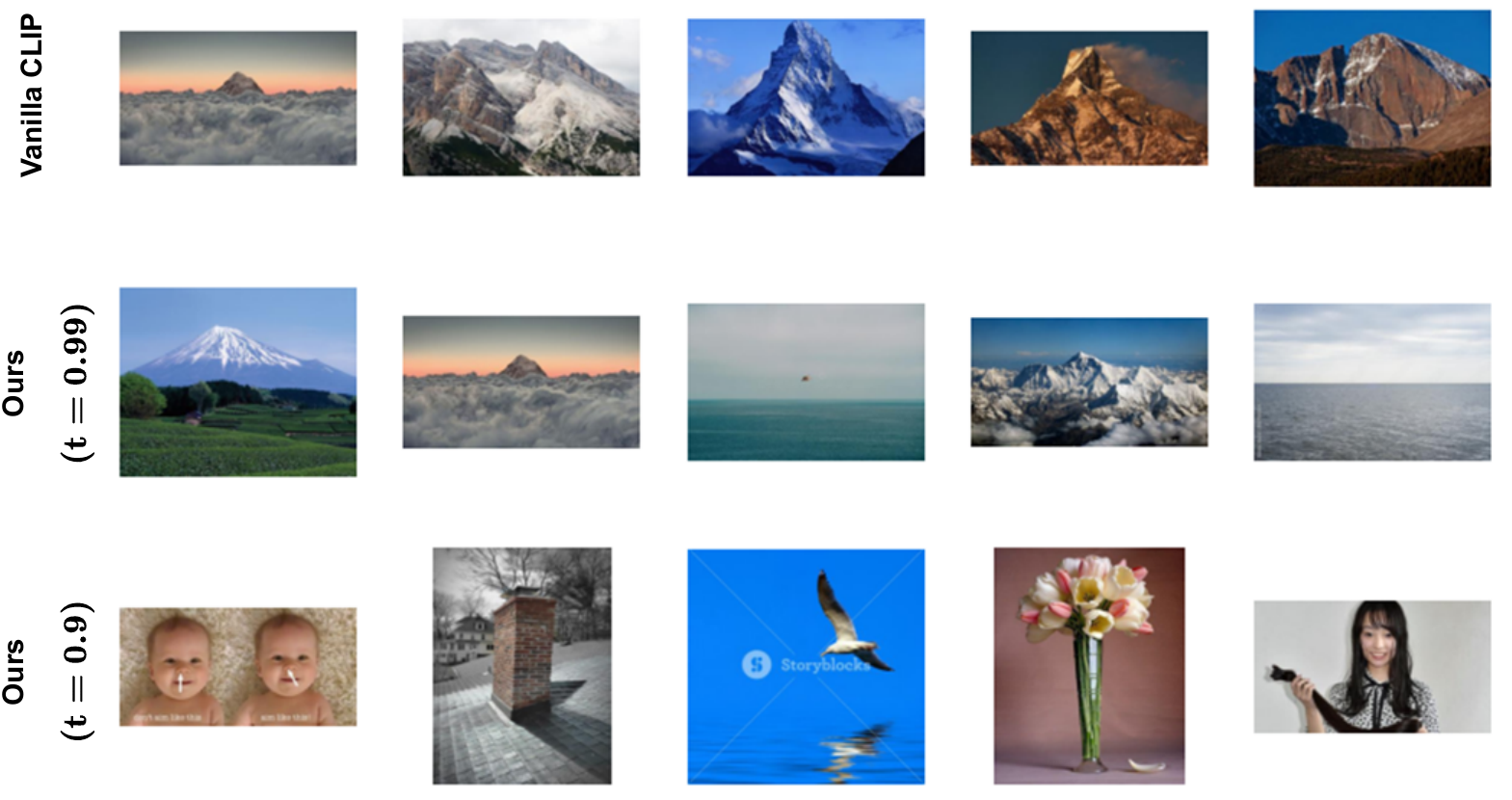}
        \caption{}
        \label{fig3:b}
    \end{subfigure}

    \caption{(a) Entropy of retrieved labels. (b) Retrieved images.}
        
\end{figure}

We compare SpaceVLM with vanilla CLIP by conducting an image-retrieval study on CIFAR-100 to evaluate both \emph{exclusion} (retrieving images outside a negated category) and \emph{diversity} among the retrieved results. We consider two settings:
\begin{inparaenum}[(i)]
    \item vanilla prompting with \texttt{"Not a photo of a <category>"},
    \item our subspace-based negation using an affirmative prompt \texttt{"This is a photo"} combined with a negation prompt \texttt{"A photo of a <category>"}.
\end{inparaenum}
Our goal is to retrieve images that do not belong to the specified category and that are semantically diverse. To quantify diversity, we compute the (Shannon) entropy~\cite{Shannon1948AMT} over the CIFAR-100 categories of the top-5 retrieved images. Figure~\ref{fig3:a} reports these entropy values: our method consistently yields higher entropy than vanilla CLIP, indicating more diverse results. Moreover, decreasing the threshold $t$ increases entropy, consistent with a larger feasible region in the embedding space. Figure~\ref{fig3:b} shows the top-5 results for the prompt \texttt{``Not a photo of a mountain''}. Our approach retrieves diverse images that are not labeled \texttt{"mountain"}, whereas vanilla CLIP often fails to account for negation. As $t$ decreases, the retrieved images further diverge from the \texttt{"mountain"} category, reflecting the expanded subspace and increased diversity. To retrieve more relevant concepts related to the negated one (e.g., \texttt{sky} to \texttt{mountain}), the threshold should be kept in the range of [0.9, 0.95]. A higher thresholds lead to the leakage of the negated concept in the retrieved image, while lower thresholds result in totally irrelevant retrieved concepts.
\vspace{-4pt}
\section{Conclusions and Limitations}
We have presented a training-free geometric framework, SpaceVLM, for modeling negation in vision–language models. It treats negation as a subspace rather than a single embedding vector, allowing joint-embedding VLMs to handle negated prompts effectively without fine-tuning. The framework depends on a lightweight language module for query decomposition, which adds minor latency but works effectively even with small models such as TinyLlama-1B. Our study focuses on joint-embedding architectures; extending the subspace formulation to sequence-conditioned models such as LLaVA~\cite{liu2023visual} is left for future work. The consistent gains across diverse backbones and tasks suggest that subspace reasoning is a natural mechanism for representing logical structure in vision–language spaces. We hope this geometric perspective will inspire further research on broader forms of logical and compositional reasoning.

\clearpage


\clearpage

{
    \small
    \bibliographystyle{ieeenat_fullname}
    \bibliography{main}

@inproceedings{alhamoud2025vision,
  title={Vision-language models do not understand negation},
  author={Alhamoud, Kumail and Alshammari, Shaden and Tian, Yonglong and Li, Guohao and Torr, Philip HS and Kim, Yoon and Ghassemi, Marzyeh},
  booktitle={Proceedings of the Computer Vision and Pattern Recognition Conference},
  pages={29612--29622},
  year={2025}
}

@article{yuksekgonul2022and,
  title={When and why vision-language models behave like bags-of-words, and what to do about it?},
  author={Yuksekgonul, Mert and Bianchi, Federico and Kalluri, Pratyusha and Jurafsky, Dan and Zou, James},
  journal={arXiv preprint arXiv:2210.01936},
  year={2022}
}

@inproceedings{radford2021learning,
  title={Learning transferable visual models from natural language supervision},
  author={Radford, Alec and Kim, Jong Wook and Hallacy, Chris and Ramesh, Aditya and Goh, Gabriel and Agarwal, Sandhini and Sastry, Girish and Askell, Amanda and Mishkin, Pamela and Clark, Jack and others},
  booktitle={International conference on machine learning},
  pages={8748--8763},
  year={2021},
  organization={PmLR}
}

@inproceedings{lin2014microsoft,
  title={Microsoft coco: Common objects in context},
  author={Lin, Tsung-Yi and Maire, Michael and Belongie, Serge and Hays, James and Perona, Pietro and Ramanan, Deva and Doll{\'a}r, Piotr and Zitnick, C Lawrence},
  booktitle={European conference on computer vision},
  pages={740--755},
  year={2014},
  organization={Springer}
}

@inproceedings{xu2016msr,
  title={Msr-vtt: A large video description dataset for bridging video and language},
  author={Xu, Jun and Mei, Tao and Yao, Ting and Rui, Yong},
  booktitle={Proceedings of the IEEE conference on computer vision and pattern recognition},
  pages={5288--5296},
  year={2016}
}

@article{pascal-voc-2007,
  title={The pascal visual object classes (voc) challenge},
  author={Everingham, Mark and Van Gool, Luc and Williams, Christopher KI and Winn, John and Zisserman, Andrew},
  journal={International journal of computer vision},
  volume={88},
  number={2},
  pages={303--338},
  year={2010},
  publisher={Springer}
}

@misc{jiang2023mistral7b,
      title={Mistral 7B}, 
      author={Albert Q. Jiang and Alexandre Sablayrolles and Arthur Mensch and Chris Bamford and Devendra Singh Chaplot and Diego de las Casas and Florian Bressand and Gianna Lengyel and Guillaume Lample and Lucile Saulnier and Lélio Renard Lavaud and Marie-Anne Lachaux and Pierre Stock and Teven Le Scao and Thibaut Lavril and Thomas Wang and Timothée Lacroix and William El Sayed},
      year={2023},
      eprint={2310.06825},
      archivePrefix={arXiv},
      primaryClass={cs.CL},
      url={https://arxiv.org/abs/2310.06825}, 
}

@inproceedings{singh2025learning,
  title={Learning the Power of “No”: Foundation Models with Negations},
  author={Singh, Jaisidh and Shrivastava, Ishaan and Vatsa, Mayank and Singh, Richa and Bharati, Aparna},
  booktitle={2025 IEEE/CVF Winter Conference on Applications of Computer Vision (WACV)},
  pages={8002--8012},
  year={2025},
  organization={IEEE}
}

@inproceedings{rombach2022high,
  title={High-resolution image synthesis with latent diffusion models},
  author={Rombach, Robin and Blattmann, Andreas and Lorenz, Dominik and Esser, Patrick and Ommer, Bj{\"o}rn},
  booktitle={Proceedings of the IEEE/CVF conference on computer vision and pattern recognition},
  pages={10684--10695},
  year={2022}
}

@inproceedings{tao2023galip,
  title={Galip: Generative adversarial clips for text-to-image synthesis},
  author={Tao, Ming and Bao, Bing-Kun and Tang, Hao and Xu, Changsheng},
  booktitle={Proceedings of the IEEE/CVF conference on computer vision and pattern recognition},
  pages={14214--14223},
  year={2023}
}

@article{park2025know,
  title={Know" No''Better: A Data-Driven Approach for Enhancing Negation Awareness in CLIP},
  author={Park, Junsung and Lee, Jungbeom and Song, Jongyoon and Yu, Sangwon and Jung, Dahuin and Yoon, Sungroh},
  journal={arXiv preprint arXiv:2501.10913},
  year={2025}
}

@article{team2025gemma,
  title={Gemma 3 technical report},
  author={Team, Gemma and Kamath, Aishwarya and Ferret, Johan and Pathak, Shreya and Vieillard, Nino and Merhej, Ramona and Perrin, Sarah and Matejovicova, Tatiana and Ram{\'e}, Alexandre and Rivi{\`e}re, Morgane and others},
  journal={arXiv preprint arXiv:2503.19786},
  year={2025}
}

@inproceedings{kamath-etal-2023-text,
    title = "Text encoders bottleneck compositionality in contrastive vision-language models",
    author = "Kamath, Amita  and
      Hessel, Jack  and
      Chang, Kai-Wei",
    editor = "Bouamor, Houda  and
      Pino, Juan  and
      Bali, Kalika",
    booktitle = "Proceedings of the 2023 Conference on Empirical Methods in Natural Language Processing",
    month = dec,
    year = "2023",
    address = "Singapore",
    publisher = "Association for Computational Linguistics",
    url = "https://aclanthology.org/2023.emnlp-main.301/",
    doi = "10.18653/v1/2023.emnlp-main.301",
    pages = "4933--4944"
}

@article{lewis2022does,
  title={Does clip bind concepts? probing compositionality in large image models},
  author={Lewis, Martha and Nayak, Nihal V and Yu, Peilin and Yu, Qinan and Merullo, Jack and Bach, Stephen H and Pavlick, Ellie},
  journal={arXiv preprint arXiv:2212.10537},
  year={2022}
}

@article{zhou2025logic,
  title={Logic Unseen: Revealing the Logical Blindspots of Vision-Language Models},
  author={Zhou, Yuchen and Tang, Jiayu and Yang, Shuo and Xiao, Xiaoyan and Dai, Yuqin and Yang, Wenhao and Gou, Chao and Xia, Xiaobo and Chua, Tat-Seng},
  journal={arXiv preprint arXiv:2508.11317},
  year={2025}
}

@article{rasekh2025multi,
  title={Multi-Rationale Explainable Object Recognition via Contrastive Conditional Inference},
  author={Rasekh, Ali and Ranjbar, Sepehr Kazemi and Gottschalk, Simon},
  journal={arXiv preprint arXiv:2508.14280},
  year={2025}
}

@inproceedings{li2022blip,
  title={Blip: Bootstrapping language-image pre-training for unified vision-language understanding and generation},
  author={Li, Junnan and Li, Dongxu and Xiong, Caiming and Hoi, Steven},
  booktitle={International conference on machine learning},
  pages={12888--12900},
  year={2022},
  organization={PMLR}
}

@inproceedings{zhai2023sigmoid,
  title={Sigmoid loss for language image pre-training},
  author={Zhai, Xiaohua and Mustafa, Basil and Kolesnikov, Alexander and Beyer, Lucas},
  booktitle={Proceedings of the IEEE/CVF international conference on computer vision},
  pages={11975--11986},
  year={2023}
}

@article{zhang2024vision,
  title={Vision-language models for vision tasks: A survey},
  author={Zhang, Jingyi and Huang, Jiaxing and Jin, Sheng and Lu, Shijian},
  journal={IEEE transactions on pattern analysis and machine intelligence},
  volume={46},
  number={8},
  pages={5625--5644},
  year={2024},
  publisher={IEEE}
}

@article{zhao2023clip,
  title={Clip in medical imaging: A comprehensive survey},
  author={Zhao, Zihao and Liu, Yuxiao and Wu, Han and Wang, Mei and Li, Yonghao and Wang, Sheng and Teng, Lin and Liu, Disheng and Cui, Zhiming and Wang, Qian and others},
  journal={arXiv preprint arXiv:2312.07353},
  year={2023}
}

@inproceedings{klemmer2025satclip,
  title={Satclip: Global, general-purpose location embeddings with satellite imagery},
  author={Klemmer, Konstantin and Rolf, Esther and Robinson, Caleb and Mackey, Lester and Ru{\ss}wurm, Marc},
  booktitle={Proceedings of the AAAI Conference on Artificial Intelligence},
  volume={39},
  number={4},
  pages={4347--4355},
  year={2025}
}

@article{luo2021clip4clip,
  title={Clip4clip: An empirical study of clip for end to end video clip retrieval},
  author={Luo, Huaishao and Ji, Lei and Zhong, Ming and Chen, Yang and Lei, Wen and Duan, Nan and Li, Tianrui},
  journal={arXiv preprint arXiv:2104.08860},
  year={2021}
}

@article{alpay2023multimodal,
  title={Multimodal video retrieval with CLIP: a user study},
  author={Alpay, Tayfun and Magg, Sven and Broze, Philipp and Speck, Daniel},
  journal={Information Retrieval Journal},
  volume={26},
  number={1},
  pages={6},
  year={2023},
  publisher={Springer}
}

@inproceedings{lulf2024clip,
  title={Clip-branches: Interactive fine-tuning for text-image retrieval},
  author={L{\"u}lf, Christian and Lima Martins, Denis Mayr and Vaz Salles, Marcos Antonio and Zhou, Yongluan and Gieseke, Fabian},
  booktitle={Proceedings of the 47th International ACM SIGIR Conference on Research and Development in Information Retrieval},
  pages={2719--2723},
  year={2024}
}

@article{liu2023visual,
  title={Visual instruction tuning},
  author={Liu, Haotian and Li, Chunyuan and Wu, Qingyang and Lee, Yong Jae},
  journal={Advances in neural information processing systems},
  volume={36},
  pages={34892--34916},
  year={2023}
}

@article{wang2022git,
  title={Git: A generative image-to-text transformer for vision and language},
  author={Wang, Jianfeng and Yang, Zhengyuan and Hu, Xiaowei and Li, Linjie and Lin, Kevin and Gan, Zhe and Liu, Zicheng and Liu, Ce and Wang, Lijuan},
  journal={arXiv preprint arXiv:2205.14100},
  year={2022}
}

@inproceedings{changpinyo2021conceptual,
  title={Conceptual 12m: Pushing web-scale image-text pre-training to recognize long-tail visual concepts},
  author={Changpinyo, Soravit and Sharma, Piyush and Ding, Nan and Soricut, Radu},
  booktitle={Proceedings of the IEEE/CVF conference on computer vision and pattern recognition},
  pages={3558--3568},
  year={2021}
}

@article{lipman2022flow,
  title={Flow matching for generative modeling},
  author={Lipman, Yaron and Chen, Ricky TQ and Ben-Hamu, Heli and Nickel, Maximilian and Le, Matt},
  journal={arXiv preprint arXiv:2210.02747},
  year={2022}
}

@inproceedings{irvin2019chexpert,
  title={Chexpert: A large chest radiograph dataset with uncertainty labels and expert comparison},
  author={Irvin, Jeremy and Rajpurkar, Pranav and Ko, Michael and Yu, Yifan and Ciurea-Ilcus, Silviana and Chute, Chris and Marklund, Henrik and Haghgoo, Behzad and Ball, Robyn and Shpanskaya, Katie and others},
  booktitle={Proceedings of the AAAI conference on artificial intelligence},
  volume={33},
  number={01},
  pages={590--597},
  year={2019}
}

@article{zhang2023biomedclip,
  title={Biomedclip: a multimodal biomedical foundation model pretrained from fifteen million scientific image-text pairs},
  author={Zhang, Sheng and Xu, Yanbo and Usuyama, Naoto and Xu, Hanwen and Bagga, Jaspreet and Tinn, Robert and Preston, Sam and Rao, Rajesh and Wei, Mu and Valluri, Naveen and others},
  journal={arXiv preprint arXiv:2303.00915},
  year={2023}
}

@article{lu2024visual,
  title={A visual-language foundation model for computational pathology},
  author={Lu, Ming Y and Chen, Bowen and Williamson, Drew FK and Chen, Richard J and Liang, Ivy and Ding, Tong and Jaume, Guillaume and Odintsov, Igor and Le, Long Phi and Gerber, Georg and others},
  journal={Nature medicine},
  volume={30},
  number={3},
  pages={863--874},
  year={2024},
  publisher={Nature Publishing Group US New York}
}

@misc{allal2024SmolLM,
  title={SmolLM - blazingly fast and remarkably powerful},
  author={Loubna Ben Allal and Anton Lozhkov and Elie Bakouch and Leandro von Werra and Thomas Wolf},
  year={2024},
}

@article{zhang2024TinyLlama,
  title        = {TinyLlama: An Open-Source Small Language Model},
  author       = {Peiyuan Zhang and Guangtao Zeng and Tianduo Wang and Wei Lu},
  journal      = {arXiv preprint arXiv:2401.02385},
  year         = {2024},
  url          = {https://arxiv.org/abs/2401.02385}
}

@misc{qwen2.5,
  title = {Qwen2.5: A Party of Foundation Models},
  url   = {https://qwenlm.github.io/blog/qwen2.5/},
  author= {Qwen Team},
  month = {September},
  year   = {2024}
}

@misc{mistral7binstruct_v0.3,
  title        = {Mistral-7B-Instruct-v0.3},
  author       = {Mistral AI Team},
  year         = {2024},
  howpublished = {\url{https://huggingface.co/mistralai/Mistral-7B-Instruct-v0.3}},
  note         = {Instruct-fine­tuned version of Mistral-7B v0.3}
}

@article{
lian2024llmgrounded,
title={{LLM}-grounded Diffusion: Enhancing Prompt Understanding of Text-to-Image Diffusion Models with Large Language Models},
author={Long Lian and Boyi Li and Adam Yala and Trevor Darrell},
journal={Transactions on Machine Learning Research},
issn={2835-8856},
year={2024},
url={https://openreview.net/forum?id=hFALpTb4fR},
note={Featured Certification}
}

@inproceedings{wu2024self,
  title={Self-correcting llm-controlled diffusion models},
  author={Wu, Tsung-Han and Lian, Long and Gonzalez, Joseph E and Li, Boyi and Darrell, Trevor},
  booktitle={Proceedings of the IEEE/CVF Conference on Computer Vision and Pattern Recognition},
  pages={6327--6336},
  year={2024}
}

@article{kang2025clip,
  title={Is CLIP ideal? No. Can we fix it? Yes!},
  author={Kang, Raphi and Song, Yue and Gkioxari, Georgia and Perona, Pietro},
  journal={ICCV},
  year={2025}
}

@inproceedings{fini2025multimodal,
  title={Multimodal autoregressive pre-training of large vision encoders},
  author={Fini, Enrico and Shukor, Mustafa and Li, Xiujun and Dufter, Philipp and Klein, Michal and Haldimann, David and Aitharaju, Sai and da Costa, Victor G Turrisi and B{\'e}thune, Louis and Gan, Zhe and others},
  booktitle={Proceedings of the Computer Vision and Pattern Recognition Conference},
  pages={9641--9654},
  year={2025}
}

@inproceedings{zhai2022lit,
  title={Lit: Zero-shot transfer with locked-image text tuning},
  author={Zhai, Xiaohua and Wang, Xiao and Mustafa, Basil and Steiner, Andreas and Keysers, Daniel and Kolesnikov, Alexander and Beyer, Lucas},
  booktitle={Proceedings of the IEEE/CVF conference on computer vision and pattern recognition},
  pages={18123--18133},
  year={2022}
}

@article{tschannen2025siglip,
  title={Siglip 2: Multilingual vision-language encoders with improved semantic understanding, localization, and dense features},
  author={Tschannen, Michael and Gritsenko, Alexey and Wang, Xiao and Naeem, Muhammad Ferjad and Alabdulmohsin, Ibrahim and Parthasarathy, Nikhil and Evans, Talfan and Beyer, Lucas and Xia, Ye and Mustafa, Basil and others},
  journal={arXiv preprint arXiv:2502.14786},
  year={2025}
}

@article{zhao2025quantifying,
  title={Quantifying Structure in CLIP Embeddings: A Statistical Framework for Concept Interpretation},
  author={Zhao, Jitian and Li, Chenghui and Sala, Frederic and Rohe, Karl},
  journal={arXiv preprint arXiv:2506.13831},
  year={2025}
}

@inproceedings{ma2023crepe,
  title={Crepe: Can vision-language foundation models reason compositionally?},
  author={Ma, Zixian and Hong, Jerry and Gul, Mustafa Omer and Gandhi, Mona and Gao, Irena and Krishna, Ranjay},
  booktitle={Proceedings of the IEEE/CVF Conference on Computer Vision and Pattern Recognition},
  pages={10910--10921},
  year={2023}
}

@article{li2024genaibench,
  title={GenAI-Bench: Evaluating and Improving Compositional Text-to-Visual Generation},
  author={Baiqi Li and Zhiqiu Lin and Deepak Pathak and Jiayao Li and Yixin Fei and Kewen Wu and Tiffany Ling and Xide Xia and Pengchuan Zhang and Graham Neubig and Deva Ramanan},
  journal={ArXiv},
  year={2024},
  volume={abs/2406.13743},
  url={https://api.semanticscholar.org/CorpusID:270619531}
}

@inproceedings{li2025exploring,
  title={Exploring how generative mllms perceive more than clip with the same vision encoder},
  author={Li, Siting and Koh, Pang Wei and Du, Simon Shaolei},
  booktitle={Proceedings of the 63rd Annual Meeting of the Association for Computational Linguistics (Volume 1: Long Papers)},
  pages={10101--10119},
  year={2025}
}

@article{Bhalla2024InterpretingCW,
  title={Interpreting CLIP with Sparse Linear Concept Embeddings (SpLiCE)},
  author={Usha Bhalla and Alexander X. Oesterling and Suraj Srinivas and Fl{\'a}vio du Pin Calmon and Himabindu Lakkaraju},
  journal={ArXiv},
  year={2024},
  volume={abs/2402.10376},
  url={https://api.semanticscholar.org/CorpusID:267740469}
}

@article{Shannon1948AMT,
  title={A mathematical theory of communication},
  author={Claude E. Shannon},
  journal={Bell Syst. Tech. J.},
  year={1948},
  volume={27},
  pages={623-656},
  url={https://api.semanticscholar.org/CorpusID:55379485}
}
}


\end{document}